\documentclass{article}

\PassOptionsToPackage{numbers, compress}{natbib}
\usepackage{booktabs}
\usepackage[preprint]{neurips_2026}

\usepackage[utf8]{inputenc} 
\usepackage[T1]{fontenc}    
\usepackage{hyperref}       
\usepackage{url}            
\usepackage{booktabs}       
\usepackage{amsfonts}       
\usepackage{nicefrac}       
\usepackage{microtype}      
\usepackage{xcolor}         
\usepackage{amssymb}
\usepackage{adjustbox}
\usepackage{booktabs}
\usepackage{amssymb}
\usepackage{adjustbox}
\usepackage{float}
\usepackage{algorithm}
\usepackage{algpseudocode}
\usepackage{graphicx} 
\usepackage{amsmath, amssymb}    
\usepackage{graphicx}           
\usepackage{xspace}
\usepackage{algorithm}          
\usepackage{algpseudocode}      
\usepackage{booktabs}
\usepackage{multirow}
\usepackage[most]{tcolorbox}
\usepackage[table]{xcolor}
\usepackage{colortbl}
\definecolor{bestcol}{RGB}{245,230,230}
\newtcolorbox{takeawaybox}{
  colback=blue!8,
  colframe=black!25,
  arc=4pt,
  boxrule=0.4pt,
  left=6pt,
  right=6pt,
  top=4pt,
  bottom=4pt,
  before skip=6pt,
  after skip=6pt
}
\definecolor{bestrow}{RGB}{245,230,230} %

\title{Breaking the Bubble: Asynchronous Pipeline Parallel Training with Bounded Weight Inconsistency}
 
\author{%
  Itay Elam \\
  Department of Computer Science\\
  Technion - Israel Institute of Technology\\
  \texttt{itayelam@gmail.com} \\
  \And
  Eliron Rahimi \\
  Department of Computer Science\\
  Technion - Israel Institute of Technology\\
  \texttt{elironrahimiacademy@gmail.com} \\
  \And
  Avi Mendelson \\
  Department of Computer Science\\
  Technion - Israel Institute of Technology\\
  \texttt{mendlson@technion.ac.il} \\
  \And
  Chaim Baskin \\
  School of Electrical and Computer Engineering\\
  Ben-Gurion University of the Negev\\
  \texttt{chaimbaskin@bgu.ac.il} \\
}

\begin{document}
\maketitle

\newcommand{\method}{\textsc{PACI}\xspace}

\begin{abstract}
Pipeline parallelism is essential for training large neural networks, but existing schedules trade off throughput, memory, and optimization consistency. Synchronous pipelines preserve forward/backward weight consistency but suffer from bubbles; asynchronous pipelines remove bubbles but introduce weight-version mismatch, typically requiring weight stashing, prediction, or correction mechanisms. We introduce \method{} (\textbf{P}ipeline \textbf{A}synchronous training with \textbf{C}ontrolled \textbf{I}nconsistency), a bubble-free asynchronous pipeline method that bounds forward/backward version drift without weight stashing, prediction, additional parameter copies, or global synchronization. The key idea is to use local gradient accumulation as a version-control mechanism: by slowing parameter-version evolution relative to pipeline delay, \method{} limits the number of optimizer updates crossed by any micro-batch while preserving steady-state utilization. In GPT-style language-model pretraining, \method{} matches the stability and final perplexity of synchronous 1F1B-flush, retains the same peak memory footprint, achieves fully utilized pipeline throughput, and improves training time-to-accuracy by up to $1.69\times$ over the fastest flush baseline. These results show that forward/backward inconsistency need not be eliminated: when explicitly bounded, it can be safely traded for substantial efficiency gains. The code is available at \href{https://github.com/ItayElam/PACI}{ \raisebox{-0.15\height}{ \includegraphics[height=1em]{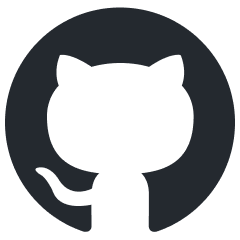} } \texttt{ItayElam/PACI} }.

\end{abstract}
\section{Introduction}
\label{sec:intro}

The rapid scaling of deep neural networks, and in particular transformer \cite{vaswani2017attention} based large language models (LLMs), has fundamentally reshaped modern machine learning systems~\cite{kaplan2020scaling, hoffmann2022training, brown2020language}. Training large neural networks efficiently requires distributing both computation and memory across accelerators~\cite{shoeybi2019megatron, huang2019gpipe}. Pipeline parallelism is a central tool for this purpose: it partitions the model across devices and overlaps computation across stages. However, pipeline-parallel training exposes a persistent trade-off between \emph{hardware utilization}, \emph{training consistency}, and \emph{memory efficiency}. Synchronous schedules preserve consistency but incur pipeline bubbles, while asynchronous schedules remove bubbles but introduce forward/backward weight-version inconsistency, where the backward pass of a micro-batch may use a later parameter version than its forward pass. Existing approaches typically attempt to mitigate this inconsistency through weight stashing or equivalent \cite{harlap2018pipedream, hosoki2024ashpipe, narayanan2021memory, yang2021pipemare} or prediction, \cite{chen2018efficient, guan2019xpipe, guan2025pipeoptim, ajanthan2025nesterov} trading off memory, computation, or system complexity.
In this work, we ask a different question: rather than attempting to eliminate forward/backward inconsistency entirely, can we make it small enough to tolerate from the source while retaining the efficiency of asynchronous execution? Our answer is \method{}, an asynchronous 1F1B pipeline schedule that controls inconsistency through micro-batch gradient accumulation. In addition to how standard gradient accumulation is used, to increase the effective batch size, \method{} uses accumulation as a local version-control mechanism: it slows parameter-version advancement relative to the bounded number of unresolved forwards in an asynchronous pipeline. The key observation is that inconsistency is governed not by asynchrony alone, but by the number of optimizer updates that occur between a micro-batch's forward and backward passes. By slowing parameter-version advancement, \method{} also reduces weight staleness, measured in parameter-version steps, between upstream and downstream stages. By accumulating gradients locally and updating less frequently, \method{} bounds inconsistency without weight stashing, prediction, or synchronization, yielding a previously unattained operating point: \textbf{zero pipeline bubbles, zero additional weight memory, and low bounded inconsistency.}
This perspective changes the role of micro-batching. In synchronous pipelines, increasing the number of micro-batches is needed to reduce bubble overhead, but due to global batch size and kernel efficiency constraints, the pipeline often remains underutilized. In \method{}, accumulation instead directly controls version drift and thus, \method{} separates the systems goal of high utilization from the optimization goal of low inconsistency.
Empirically, we show that this low-inconsistency regime where bubble overhead is non-negligible is sufficient for stable training and provides substantial wall-clock gains. On GPT-2 Medium pretraining from scratch on OpenWebText, \method{} closely matches the loss dynamics and final validation perplexity of synchronous 1F1B-flush, while reducing run-to-run variability. The removed bubbles translate directly into faster convergence: \method{} reaches target perplexities earlier and reduces end-to-end runtime by up to $1.69\times$ at batch size 128 and $1.41\times$ at batch size 256 compared with the fastest flush baselines. We further show that synchronous 1F1B throughput follows the predicted bubble-efficiency scaling, while \method{} achieves the corresponding fully-utilized throughput at the same memory footprint.

Our contributions are:
\begin{itemize}    
\item \textbf{Bounded inconsistency without stashing.}
    We show that local gradient accumulation can serve as a tunable parameter-version control mechanism, bounding forward/backward inconsistency without extra weight memory, prediction, or synchronization.
    \item \textbf{Improved training time-to-accuracy with preserved quality.}
    We demonstrate stable pretraining with comparable final perplexity while achieving up to a $1.69\times$ speedup over the fastest flush baseline.
    \item \textbf{Theory-matched throughput analysis.}
    We show that the throughput gap between synchronous 1F1B-flush and \method{} is explained by pipeline bubble efficiency, and that increasing micro-batches in flush trades bubble reduction against kernel efficiency.

\end{itemize}

\begin{figure}[t]
    \centering
    \includegraphics[width=0.9\linewidth]{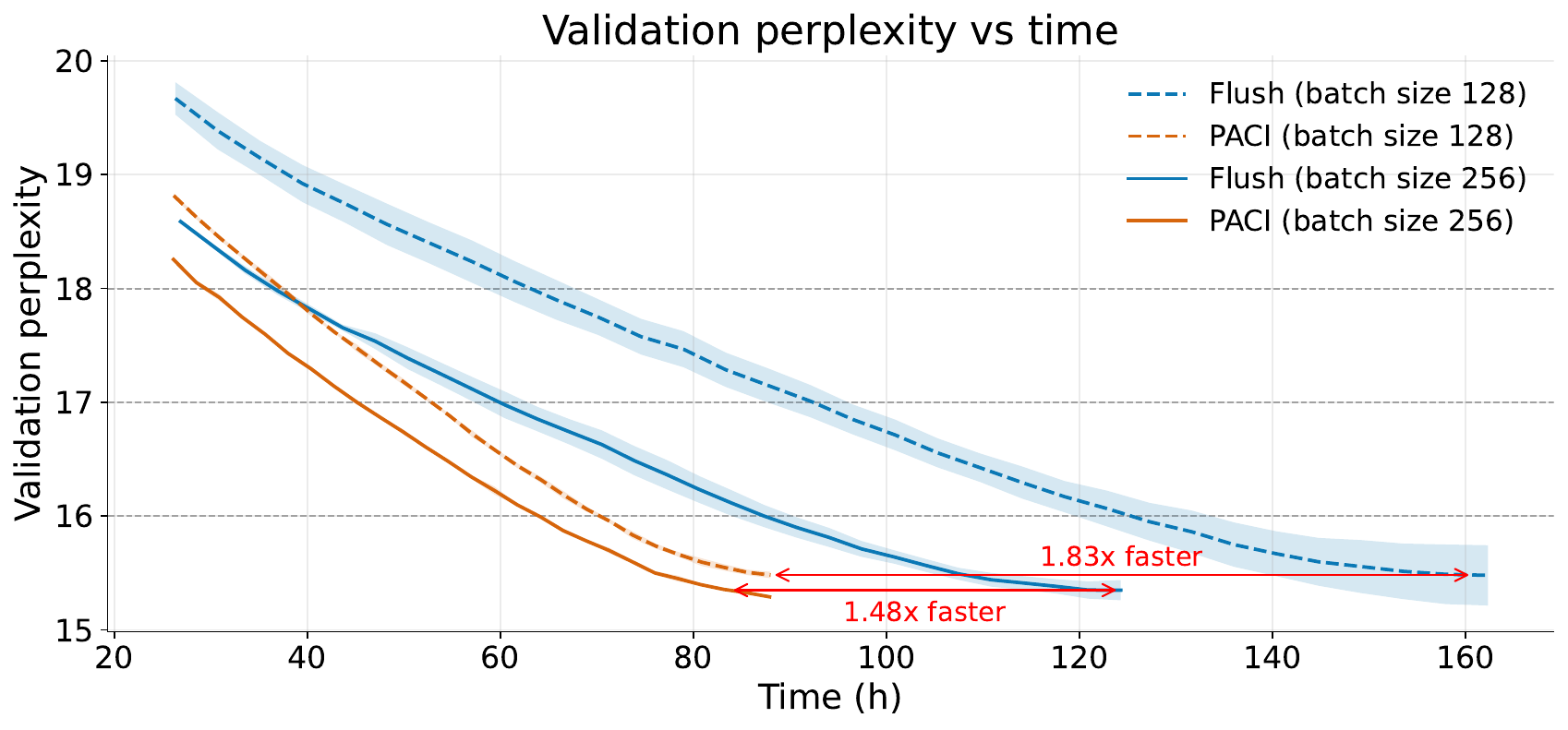}
    \caption{
    Validation perplexity versus wall-clock time. 
    \method{} reaches the same perplexity levels earlier than 1F1B-flush, showing that removing pipeline bubbles while controlling weight inconsistency translates into improved training time-to-accuracy rather than only higher raw throughput.
    }
    \label{fig:perplexity_vs_time}
\end{figure}

\section{Related work}
\label{sec:related}

\begin{table*}[t]
\caption{
Qualitative trade-offs among representative pipeline-parallel schedules.
See Appendix~\ref{app:app_related} for a more detailed table.
}

\centering
\label{tab:pipeline_tradeoffs}
\footnotesize
\setlength{\tabcolsep}{4pt}
\renewcommand{\arraystretch}{1.12}
\begin{tabular}{lccccccc>{\columncolor{bestcol}}c}
\toprule
\textbf{Property}
& \textbf{Flush}
& \textbf{1F1B-I}
& \textbf{ZB-2p}
& \textbf{Na\"ive 1F1B}
& \textbf{2BW}
& \textbf{PipeMare}
& \textbf{PipeOptim}
& \textbf{\method{} (Ours)} \\
\midrule
Execution
& Sync.
& Sync.
& Sync.
& Async.
& Async.
& Async.
& Async.
& Async. \\

Extra memory
& 0
& +
& ++
& 0
& ++
& ++
& +/++
& \textbf{0} \\

Pipeline bubble
& High
& Reduced
& Near-0
& 0
& 0
& 0
& 0
& \textbf{0} \\

F/B inconsistency
& 0
& 0
& 0
& High
& 0
& High
& Approx.
& \textbf{Low (Bounded)} \\
\bottomrule
\end{tabular}
\end{table*}

\paragraph{Synchronous pipeline parallelism}
Synchronous PP methods preserve standard mini-batch training semantics: all micro-batches contributing to an optimizer step are evaluated under a consistent parameter version, and the update is applied only after all respective backwards are completed~\cite{narayanan2021memory, huang2019gpipe}. This eliminates forward/backward weight-version inconsistency, but requires flushing or coordinated execution, which introduces idle time and reduces utilization. GPipe~\cite{huang2019gpipe} is the canonical synchronous schedule, executing all forward passes before backward propagation and draining the pipeline before each update. Later methods reduce bubbles or memory consumption through improved partitioning and scheduling, including DAPPLE~\cite{fan2021dapple}, PipeDream-Flush~\cite{narayanan2021memory}, Megatron-LM interleaved 1F1B~\cite{narayanan2021efficient}, Chimera~\cite{li2021chimera}, and Seq1F1B~\cite{sun2024seq1f1b}. Zero-Bubble Pipeline Parallelism~\cite{qi2023zero} further reduces idle time by decomposing backward computation and scheduling work into bubbles, with variants that trade non-trivial balanced computation time assumption or additional memory for lower bubble overhead. Overall, follow-up work on synchronous PP has substantially improved pipeline utilization, but its core limitation remains: consistency is obtained through coordination, which inherently exposes synchronization-induced idle time. Existing schedules mitigate this overhead to varying degrees; more aggressive bubble-hiding approaches typically do so by incurring additional memory, scheduling complexity, or assumptions about balanced computation. 

\paragraph{Asynchronous pipeline parallelism.}
Asynchronous PP removes global synchronization barriers at the cost of forward/backward weight-version inconsistency: for the same micro-batch, a stage may use one parameter version during the forward pass and a later version during the backward pass. This mismatch is distinct from global weight staleness, which measures lag relative to a serial or globally synchronized parameter trajectory, but both are expressed in parameter-version steps and can affect convergence. A common way to control inconsistency is to introduce additional parameter-version state. PipeDream~\cite{harlap2018pipedream} uses weight stashing so each backward pass reuses the corresponding forward-pass weights, eliminating forward/backward inconsistency at substantial memory cost. PipeDream-2BW~\cite{narayanan2021memory} reduces stored versions through double-buffering, but still uses additional parameter copies. PipeMare~\cite{yang2021pipemare} tolerates asynchronous delay and forward/backward mismatch via learning-rate rescheduling and discrepancy correction, requiring additional weight-sized weight velocity state for stable convergence. AshPipe~\cite{hosoki2024ashpipe} uses stage-aware recomputation to reduce memory pressure caused by multiple weight versions, while preserving forward/backward weight-version consistency via weight stashing. 
A second direction uses prediction or correction mechanisms. SpecTrain~\cite{chen2018efficient}, XPipe~\cite{guan2019xpipe}, Nesterov-based methods~\cite{ajanthan2025nesterov}, and PipeOptim~\cite{guan2025pipeoptim} predict or approximate future parameter versions during the forward pass, aiming to match the weights that will be present when backward executes. These methods reduce the impact of inconsistency while preserving asynchronous throughput, but introduce additional computation, optimizer-specific assumptions, memory overhead, and non-standard computation semantics. 
Recently, a third direction introduced by AMDP~\cite{chenamdp}, attempts to control the inconsistency through scheduling, limiting the read-ahead of each asynchronous pipeline. Building upon Chimera~\cite{li2021chimera}, they recover utilization by running multiple concurrent directional pipelines. This reduces mismatch at the cost of a more complex multi-pipeline schedule, multiple different stages per GPU, and gradient synchronization across replicated logical stages, with accumulation used to amortize synchronization and cap inconsistency to one.
Overall, asynchronous PP methods typically control inconsistency by storing additional parameter-version state, predicting future weights, or constraining the schedule itself. These choices trade asynchronous utilization for extra memory, auxiliary computation, optimizer-specific assumptions, non-standard computational semantics, or more complex multi-pipeline coordination. In contrast, \method{} bounds forward/backward inconsistency at the source and reduces staleness, without weight stashing, extra parameter buffers, prediction, or replicated directional pipelines. Table~\ref{tab:pipeline_tradeoffs} summarizes these trade-offs; a more detailed comparison appears in Appendix~\ref{app:app_related}.
\section{Method}
\label{sec:method}

The key observation behind \method{} is that forward/backward inconsistency is governed not by asynchrony alone, but by the number of optimizer updates that occur during the pipeline delay between a micro-batch's forward and backward passes. \method{} exploits this separation by leaving the bubble-free asynchronous 1F1B schedule intact while slowing parameter-version evolution through local gradient accumulation. In this view, gradient accumulation serves as a \emph{version-control mechanism}, not merely as a batching tool. The result is a distinct operating point in the pipeline-parallel design space: zero pipeline bubbles, no additional weight memory, and an explicit tunable bound on forward/backward inconsistency. The rest of this section develops this mechanism, contrasts it with the use of micro-batching to amortize bubbles in flush-based schedules, and provide theoretical motivation using ~\cite{qi2023zero} reported result and extrapolating based on our experiments the effects of \method{} in large-scale training.

\begin{figure}[h]
    \centering
    \includegraphics[width=1\linewidth]{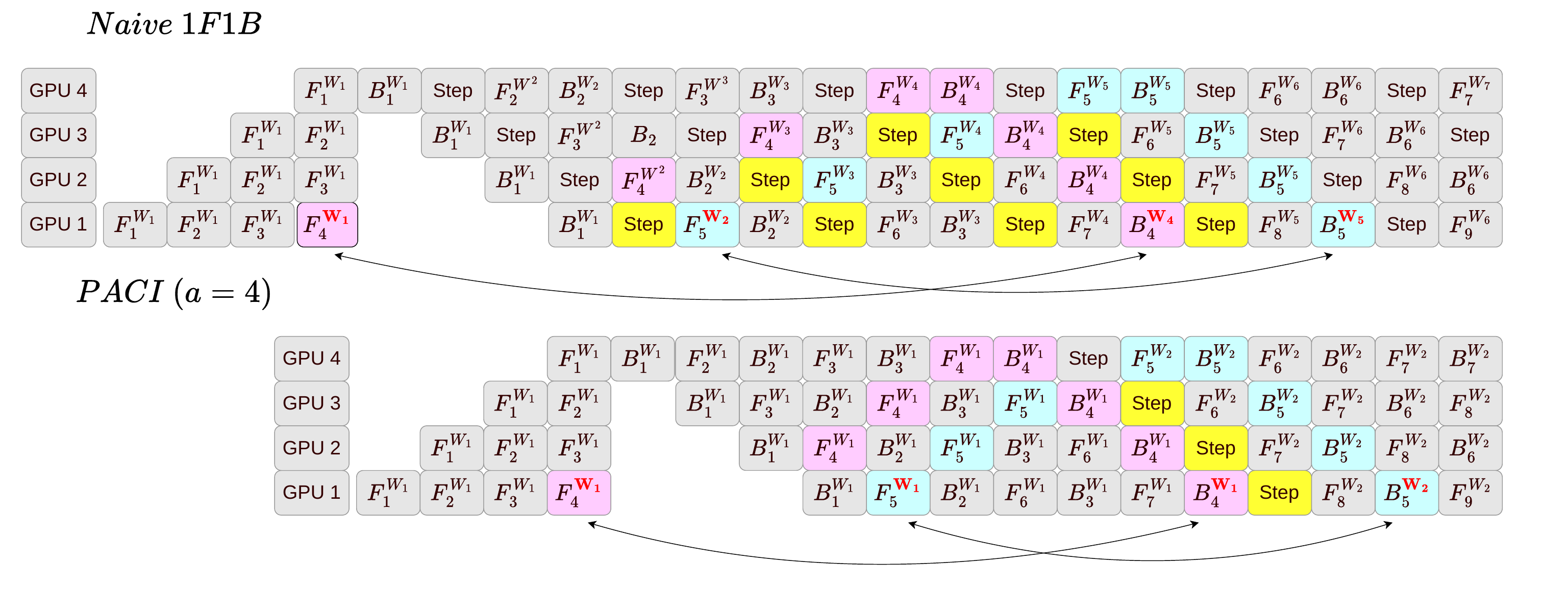}
    \caption{
    Effect of accumulation on forward/backward weight-version inconsistency. 
    With $a=1$ (top), several optimizer updates may occur between the forward and backward passes of the same micro-batch. 
    With $a=4$ (bottom), parameter versions evolve more slowly, so each micro-batch crosses fewer updates. 
    Thus, accumulation acts as a version-control mechanism for asynchronous 1F1B.
    }
    \label{fig:method_overview}
\end{figure}

\subsection{Forward/backward inconsistency as version drift}\label{sec:version_drift}

Consider an $N$-stage pipeline, where stage $i$ has parameters $\theta_i$, and let $\theta_i^{(t)}$ denote the parameters after its $t$-th local optimizer update. In asynchronous 1F1B, stages process forward and backward events as soon as their inputs arrive, without a global flush. Thus, if micro-batch $m$ is forwarded through stage $i$ using $\theta_i^{(t)}$, its backward pass may reach the same stage after $\Delta_i$ local updates, when the stage holds $\theta_i^{(t+\Delta_i)}$. We define $\Delta_i$ as the \emph{forward/backward weight-version inconsistency}: the number of parameter updates between the forward and backward computations of the same micro-batch.
The corresponding gradient uses the activation from the earlier forward pass but is evaluated at the current parameter version:
\begin{equation}
g_{m,i}
=
\delta_{m,i}
\left.
\nabla_{\theta_i} F_i\!\left(h_{m,i-1}^{(t)}; \theta_i\right)
\right|_{\theta_i = \theta_i^{(t+\Delta_i)}} .
\label{eq:gradient_inconsistency}
\end{equation}
Here $\delta_{m,i}$ denotes the downstream activation gradient. The relevant delay is therefore the number of parameter versions crossed; this is the version drift that \method{} controls.
\subsection{Controlling version drift by update frequency}
\label{sec:inconsistency_control}

\method{} modifies Naive asynchronous 1F1B by decoupling \emph{backward computation} from \emph{parameter updates}. Each stage executes backward operations as soon as their inputs arrive, but does not apply an optimizer step after every backward pass. Instead, it accumulates gradients for $a$ local backward passes and updates once per accumulation window. The accumulation factor $a$ therefore controls the rate at which parameter versions evolve relative to the pipeline delay.
To make this delay bound robust to non-uniform stage times, \method{} uses a local flow-control rule. Each stage $i$ maintains a counter $q_i$ of forward passes whose corresponding backward passes have not yet returned. The counter is incremented on each forward pass and decremented on the corresponding backward pass. A new forward is admitted only if, before admission, $q_i < N+1-i$. With the one-indexed stage convention used here, $N-i$ is the downstream pipeline depth; since $q_i$ is integer-valued and counts previously admitted unresolved forwards, the rule ensures that a newly admitted micro-batch has at most $N-i$ earlier unresolved forwards ahead of it.
This rule is not a flush or a synchronization barrier. Rather, it is a local backpressure mechanism that prevents faster upstream stages from running arbitrarily far ahead of a slower downstream stage under realistic, imperfectly balanced partitions. It bounds queue growth and activation storage while leaving the steady-state throughput determined by the bottleneck stage.
Since parameter versions advance only once every $a$ local backward passes, the bounded unresolved-forward lag implies that the number of intervening parameter updates is bounded by
\begin{equation}
\Delta_{i}
\;\le\;
\left\lceil \frac{N-i}{a} \right\rceil,
\qquad
\Delta_{\max} = \Delta_{1}
\;\le\;
\left\lceil \frac{N-1}{a} \right\rceil .
\label{eq:inconsistency_max}
\end{equation}
The same mechanism also reduces staleness: because each stage advances its local parameter version only once every $a$ backward passes, the version gap between upstream and downstream stages grows more slowly in parameter-version steps. Thus, $a$ is an explicit consistency knob: increasing $a$ reduces the number of parameter versions crossed during a micro-batch's forward/backward delay, while preserving asynchronous 1F1B execution.
Importantly, \method{} achieves this without storing old weights, predicting future weights, or introducing global synchronization. Each stage keeps a single parameter copy, and gradient accumulation reuses the standard gradient buffer already present in the optimizer; no separate parameter-version storage or predicted-weight buffer is required. The complete stage-level event rule is given in Appendix~\ref{sec:appendix_method_details}.

\subsection{Micro-batching for throughput versus consistency}
\label{sec:analysis}
Micro-batching has different leverage in synchronous and asynchronous pipelines. In synchronous 1F1B-flush, increasing $m$ improves utilization by amortizing bubbles. For an $N$-stage pipeline, the standard pipeline efficiency is~\cite{huang2019gpipe}
\begin{equation}
\eta_{\mathrm{flush}}(m,N)
=
\frac{m}{m+N-1}.
\label{eq:bubble_efficiency_flush}
\end{equation}
Interleaved 1F1B reduces the bubble term by splitting each physical stage into $V$ virtual stages. Using the bubble-size expression from~\cite{narayanan2021efficient}, the corresponding efficiency is
\begin{equation}
\eta_{\mathrm{inter}}(m,N,V)
=
\frac{m\cdot V}{m\cdot V+N-1} .
\label{eq:bubble_efficiency_interleaved}
\end{equation}

In \method{}, the same micro-batching budget is used for consistency rather than bubble amortization. With accumulation factor $m$, the worst-case forward/backward inconsistency is bounded by $\left\lceil \frac{N-1}{a} \right\rceil $ Eq~\eqref{eq:inconsistency_max}
To reach $\Delta_{\max}\le 2$, \method{} only needs $m \approx (N-1)/2$. At this point, 1F1B-flush has efficiency $1/3$, i.e. a $66\%$ bubble fraction, while interleaved 1F1B with $V=2$ has efficiency $1/2$, i.e. a $50\%$ bubble fraction. Thus, as we experimentally show, moderate micro-batching is enough to control version drift, but not enough to recover synchronous pipeline utilization. Figure~\ref{fig:inconsistency_vs_bubble} visualizes this gap.

\subsection{Projected large-scale throughput and memory}
\label{subsec:theoretical_scaling}

To assess the large-scale regime, we compare against the throughput and memory measurements reported by Zero-Bubble Pipeline Parallelism~\cite{qi2023zero}. Their experiments use up to 32 NVIDIA A100 SXM 80GB GPUs across four nodes, and report throughput for model sizes up to 28.3B parameters under multiple micro-batch counts. Importantly, these settings use relatively large numbers of micro-batches: all configurations satisfy $m\geq3N$. Thus, the comparison is not restricted to a bubble-dominated corner case; the synchronous baselines already operate in a regime where micro-batching substantially amortizes flush bubbles.
We estimate the throughput of \method{} from the reported 1F1B-flush throughput using the pipeline-efficiency model from Section~\ref{sec:analysis}. For an $N$-stage pipeline with $m$ micro-batches, synchronous 1F1B-flush efficiency scales according to Eq.~\eqref{eq:bubble_efficiency_flush}
Since \method{} preserves asynchronous 1F1B execution and does not introduce flushes or global synchronization, its ideal steady-state throughput is the fully utilized counterpart of the same pipeline. We therefore use
\begin{equation}
\widehat{T}_{PACI}
=
\frac{T_{\mathrm{1F1B\text{-}flush}}}{\eta_{\mathrm{flush}}(m,N)}
\label{eq:method_throughput_extrapolation}
\end{equation}
as a throughput proxy. This extrapolation is supported empirically by Section~\ref{subsec:throughput_scaling}, where the measured ratio between \method{} and 1F1B-flush closely follows the inverse bubble-efficiency factor.
Since \method{} adds no memory beyond 1F1B-flush, we use the measured 1F1B-flush peak memory for \method{} in the fully utilized regimes considered here, consistent with Figure~\ref{fig:max_mem_vs_num_micro}. Table~\ref{tab:theoretical_results} reports this memory estimate together with the throughput estimate from Eq.~\ref{eq:method_throughput_extrapolation}.

\begin{table*}[h]
\centering
\caption{
Theoretical throughput and peak memory comparison across model scales.
Throughput is reported as samples per GPU per second.
}
\label{tab:theoretical_results}
\scriptsize
\setlength{\tabcolsep}{3.5pt}
\renewcommand{\arraystretch}{1.08}
\begin{tabular}{l|l|ccc|ccc|ccc|ccc}
\toprule
& \textbf{Model}
& \multicolumn{3}{c|}{\textbf{1.5B}}
& \multicolumn{3}{c|}{\textbf{6.2B}}
& \multicolumn{3}{c|}{\textbf{14.6B}}
& \multicolumn{3}{c}{\textbf{28.3B}} \\
\cmidrule(lr){3-5}
\cmidrule(lr){6-8}
\cmidrule(lr){9-11}
\cmidrule(lr){12-14}
\textbf{Setup}
& \textbf{\#GPU}
& \multicolumn{3}{c|}{8}
& \multicolumn{3}{c|}{8}
& \multicolumn{3}{c|}{16}
& \multicolumn{3}{c}{32} \\
\cmidrule(lr){3-5}
\cmidrule(lr){6-8}
\cmidrule(lr){9-11}
\cmidrule(lr){12-14}
& \textbf{\#Microbatch}
& 24 & 32 & 64
& 24 & 32 & 64
& 48 & 64 & 128
& 96 & 128 & 256 \\
\midrule

\multirow{6}{*}{\shortstack{\textbf{Samples}\\\textbf{per GPU}\\\textbf{per second}}}
& \cellcolor{bestrow} \textbf{\method{}}
& \cellcolor{bestrow}\textbf{15.24} & \cellcolor{bestrow}\textbf{15.23} & \cellcolor{bestrow}\textbf{15.09}
& \cellcolor{bestrow}\textbf{4.52} & \cellcolor{bestrow}\textbf{4.51} & \cellcolor{bestrow}\textbf{4.47}
& \cellcolor{bestrow}\textbf{1.84} & \cellcolor{bestrow}\textbf{1.84} & \cellcolor{bestrow}1.83
&\cellcolor{bestrow} \textbf{1.01} & \cellcolor{bestrow}0.99 & \cellcolor{bestrow}0.99 \\
& ZB-2p
& 14.50 & 14.80 & 14.90
& 4.32 & 4.35 & 4.39
& 1.81 & 1.83 & \textbf{1.85}
& 0.99 & \textbf{1.00} & \textbf{1.00} \\
& ZB-1p
& 12.90 & 13.40 & 14.20
& 3.88 & 4.00 & 4.20
& 1.61 & 1.67 & 1.76
& 0.87 & 0.90 & 0.96 \\
& 1F1B-Flush
& 11.80 & 12.50 & 13.60
& 3.50 & 3.70 & 4.03
& 1.40 & 1.49 & 1.64
& 0.76 & 0.80 & 0.88 \\
& 1F1B-I
& 13.10 & 13.40 & 13.90
& 4.01 & 4.08 & 4.19
& 1.54 & 1.59 & 1.66
& 0.82 & 0.85 & 0.90 \\
\midrule

\multirow{5}{*}{\shortstack{\textbf{Memory}\\\textbf{(GB)}}}
& \cellcolor{bestrow}\textbf{\method{}}
& \cellcolor{bestrow}\textbf{30} & \cellcolor{bestrow}\textbf{30} & \cellcolor{bestrow}\textbf{30}
& \cellcolor{bestrow}\textbf{39} & \cellcolor{bestrow}\textbf{39} & \cellcolor{bestrow}\textbf{39}
& \cellcolor{bestrow}\textbf{32} & \cellcolor{bestrow}\textbf{32} & \cellcolor{bestrow}\textbf{32}
& \cellcolor{bestrow}\textbf{43} & \cellcolor{bestrow}\textbf{43} & \cellcolor{bestrow}\textbf{43} \\
& ZB-2p
& 59 & 59 & 59
& 70 & 70 & 70
& 51 & 51 & 51
& 74 & 74 & 74 \\
& ZB-1p
& 32 & 32 & 32
& 42 & 42 & 42
& 33 & 33 & 33
& 44 & 44 & 44 \\
& 1F1B-Flush
& \textbf{30} & \textbf{30} & \textbf{30}
& \textbf{39} & \textbf{39} & \textbf{39}
& \textbf{32} & \textbf{32} & \textbf{32}
& \textbf{43} & \textbf{43} & \textbf{43} \\
& 1F1B-I
& 40 & 40 & 40
& 48 & 48 & 48
& 39 & 39 & 39
& 58 & 58 & 58 \\
\bottomrule
\end{tabular}
\end{table*}

The resulting comparison shows the main scaling implication of \method{}: it reaches the throughput regime of ZB-2p, and in several cases exceeds it, while retaining the memory footprint of 1F1B-flush and ZB-1p. Unlike ZB-2p, this throughput does not require additional pipeline-buffer memory; unlike ZB-1p or interleaved 1F1B, it does not rely on residual bubble reduction. Instead, the speedup comes from removing flush bubbles while controlling the resulting version drift through accumulation.
Finally, the configurations in Table~\ref{tab:theoretical_results} all lie in the low-inconsistency regime. The largest inconsistency is when the pipeline has $N=32$ stages and micro-batch count is $m=96$, giving according to Eq.~\eqref{eq:inconsistency_max} $\Delta_{\max} \le \left\lceil \frac{32-1}{96} \right\rceil= 1 .$
Thus, the extrapolated large-scale gains occur in the same bounded-delay regime studied experimentally in Section~\ref{subsec:training_stability}, where \method{} achieves stable training with loss and perplexity comparable to, or better than, 1F1B-flush.

\section{Results}
\label{sec:experiments}
In this section, we conduct experiments aiming to answer these core questions:
\begin{itemize}
    \item \textbf{Q1} Are realistic ranges of forward/backward weight-version inconsistency, controlled by gradient accumulation, sufficiently low to ensure stable training, as evidenced by training loss dynamics?
    
    \item \textbf{Q2} How does \method{} compare to synchronous 1F1B-flush in terms of  training time-to-accuracy and final perplexity under a fixed token budget?

    \item \textbf{Q3} Does \method{} achieve the throughput predicted by pipeline-efficiency theory while maintaining the same memory footprint as synchronous 1F1B-flush?
\end{itemize}

\subsection{Experimental setup}
\label{subsec:experimental_setup}

We evaluate \method{} on causal language-model pretraining. Unless otherwise stated, experiments use GPT-2 Medium~\cite{radford2019language} trained from scratch on OpenWebText with sequence length 1024, AdamW~\cite{loshchilov2017decoupled}, BF16 precision~\cite{kalamkar2019study}, and a fixed token budget of 49.8B tokens. Full preprocessing, optimizer settings, and reproducibility details are provided in Appendix~\ref{sec:appendix_experimental_setup}.
All main experiments use 8-stage pipeline parallelism without data parallelism. We compare synchronous 1F1B-flush and \method{} under the same model partitioning and hardware configuration, with global batch sizes 128 and 256. For \method{}, the accumulation factor $a$ controls the forward/backward inconsistency bound in Eq.~\eqref{eq:inconsistency_max}. For 1F1B-flush, different micro-batch counts change throughput but not the optimizer trajectory under fixed batch size, data order, and partitioning; therefore, we fully train the fastest flush configuration and map the same validation trajectory to other flush runtimes using separately measured throughputs. All \method{} configurations are trained end-to-end.
Experiments run on a single 8-GPU PCIe node, Using NVIDIA RTX PRO 6000 Blackwell Max-Q GPUs with 96GB memory. We do not use activation checkpointing, ZeRO~\cite{rajbhandari2020zero}, or FSDP~\cite{zhao2023pytorch}. We report intermediate and final validation perplexity, training time-to-accuracy, global tokens/second, and peak per-device memory. Additional stability, memory, and throughput analyses are provided in Appendix~\ref{app:additional_results}.

\subsection{Low inconsistency preserves stable training}
\label{subsec:training_stability}

\begin{takeawaybox}
\textbf{Key Takeaway.} Bounded forward/backward inconsistency does not destabilize training. For $\Delta_{\max}\leq 2$, \method{} closely matches 1F1B-flush loss dynamics over 50B tokens, with comparable final loss and lower run-to-run variability.
\end{takeawaybox}

To answer Question~\textbf{Q1}, we train GPT2-M on OpenWebText for 50B tokens with three random seeds on 8 GPUs. We compare synchronous 1F1B-flush with \method{} at global batch sizes 128 and 256, varying the accumulation factor $a$. For pipeline depth $n=8$, Eq.~\eqref{eq:inconsistency_max} gives $\Delta_{\max}\leq 2$ for $a=4$ and $\Delta_{\max}\leq 1$ for $a=8$.
Figure~\ref{fig:training_loss_vs_tokens} shows that \method{} tracks the 1F1B-flush loss trajectory throughout training. The bounded inconsistency appears only as a small vertical shift in the loss curve, not as spikes, oscillations, or divergence. Additionally, this vertical shift reduces in size as training progresses as evident by table~\ref{tab:time_to_accuracy}, as training progresses the relative speedup increases as the difference in tokens to perplexity decreases. This is the central stability result: in the low-inconsistency regime, stale forward activations do not qualitatively change optimization dynamics.
Table~\ref{tab:rms_stability} confirms this quantitatively. Across both batch sizes, \method{} achieves final losses comparable to or slightly better than 1F1B-flush. At batch size 256, \method{} with $a=8$ matches the flush baseline, while \method{} with $a=16$ slightly improves the final loss. At batch size 128, \method{} with $a=8$ matches the baseline, while $a=4$ remains within a small margin. The same table also shows that \method{} reduces run-to-run variability: for example, at batch size 128, the RMS standard deviation of training loss decreases from $1.10\times10^{-2}$ for 1F1B-flush to $2.12\times10^{-3}$ for \method{} with $a=4$ and $1.81\times10^{-3}$ with $a=8$.
We observe occasional divergence or early plateau in isolated runs, but these failures are not correlated with forward/backward inconsistency: they also occur under synchronous 1F1B-flush. Thus, the observed instabilities appear attributable to inherent training variance and recipe sensitivity rather than to the bounded inconsistency introduced by \method{}.
Overall, these results show that forward/backward inconsistency up to $\Delta_{\max}\leq 2$ preserves stable training: \method{} maintains smooth convergence, comparable final loss, and lower seed-to-seed variability than synchronous 1F1B-flush.

\begin{table}[h]
\caption{
Final loss and run-to-run variability measured as RMS of standard deviation of training loss across token bins.
\method{} exhibits equal or lower final training loss with lower variability compared to 1F1B-flush.
}

\centering
\small
\begin{tabular}{lccc}
\toprule
\textbf{Method} & \textbf{Batch} & \textbf{Final Loss} $\downarrow$ & \textbf{RMS Std (Train Loss)} $\downarrow$ \\
\midrule
1F1B-Flush       & 128 & $2.79 \pm 1.94 \times 10^{-2}$ & $1.10 \times 10^{-2}$ \\
\method{} ($a=4$)  & 128 & $2.80 \pm 7.51 \times 10^{-4}$ & $2.12 \times 10^{-3}$ \\
\rowcolor{bestrow}
\method{} ($a=8$)  & 128 & $\mathbf{2.79} \pm 1.67 \times 10^{-3}$ & $\mathbf{1.81 \times 10^{-3}}$ \\
\midrule
1F1B-Flush       & 256 & $2.78 \pm 5.72 \times 10^{-3}$ & $5.30 \times 10^{-3}$ \\
\method{} ($a=8$)  & 256 & $2.78 \pm 2.90 \times 10^{-3}$ & $3.01 \times 10^{-3}$ \\
\rowcolor{bestrow}
\method{} ($a=16$) & 256 & $\mathbf{2.77} \pm 1.30 \times 10^{-3}$ & $\mathbf{1.75 \times 10^{-3}}$ \\
\bottomrule
\end{tabular}
\label{tab:rms_stability}
\end{table}

\subsection{\method{} accelerates training time-to-accuracy}\label{sec:throughput}

\begin{takeawaybox}
\textbf{Key Takeaway.} \method{} turns stable low-inconsistency training into substantial wall-clock gains: it achieves up to $2.04\times$ speedup over matched 1F1B-flush configurations, and remains up to $1.69\times$ faster even against the best-tuned flush baseline, with negligible final-perplexity differences.
\end{takeawaybox}

To answer Question~\textbf{Q2}, we compare validation perplexity as a function of wall-clock time under a fixed 49.8B-token budget. All methods use identical data order, evaluation method, and total tokens. For 1F1B-flush, we include multiple micro-batch configurations and compare against both the matched configuration and the fastest flush baseline for each batch size, giving the synchronous baseline its strongest operating point.
Figure~\ref{fig:perplexity_vs_time} shows the central result: \method{} reaches the same perplexity levels earlier than 1F1B-flush. This is not just a throughput improvement in isolation; it directly translates into faster convergence in wall-clock time. At batch size 128, \method{} reaches its final perplexity $1.84\times$ faster than the corresponding flush baseline, while also achieving a $1.69\times$ speedup over the fastest flush configuration at comparable final perplexity. Thus, in the regime where 1F1B-flush is most bubble-limited, \method{} turns the removed bubbles into immediate training time-to-accuracy improvements.
Table~\ref{tab:final_ppl_time} reports two speedups. 
Speedup$_{\mathrm{match}}$ isolates the scheduling effect by comparing \method{} to 1F1B-flush at the same number of micro-batches, while Speedup$_{\mathrm{best}}$ compares against the fastest flush configuration for the same batch size. 
The matched comparison shows the direct benefit of eliminating pipeline bubbles, reaching up to $2.04\times$ speedup at batch size 128 and $1.51\times$ at batch size 256.
The best-flush comparison is more conservative: even against the strongest synchronous baseline, \method{} still reduces end-to-end runtime by $1.69\times$ at batch size 128 and $1.41\times$ at batch size 256, with comparable or better final perplexity.
These results establish the practical payoff of bounded inconsistency. Section~\ref{subsec:training_stability} showed that low inconsistency does not destabilize training; here we show that the same regime substantially improves wall-clock efficiency. \method{} therefore provides the desired tradeoff: it preserves the optimization behavior of synchronous 1F1B-flush while eliminating enough pipeline idle time to reach useful model quality significantly sooner.

\begin{table}[t]
\centering
\caption{
Final runtime and validation perplexity after a fixed 49.8B-token budget.
Speedup$_{\mathrm{match}}$ compares \method{} against 1F1B-flush with the same number of micro-batches.
Speedup$_{\mathrm{best}}$ compares each configuration against the fastest 1F1B-flush baseline for the same batch size.
}
\label{tab:final_ppl_time}
\footnotesize
\begin{tabular}{cccrccc}
\toprule
Batch & Method & Num. micro & Runtime $\downarrow$ & Final PPL $\downarrow$
& Speedup$_{\mathrm{match}} \uparrow$ & Speedup$_{\mathrm{best}} \uparrow$ \\
\midrule
128 & Flush & 4  & $227.62$ & $\mathbf{15.480} \pm 0.265$ & $1.00\times$ & $0.65\times$ \\
128 & Flush & 8  & $162.33$ & $\mathbf{15.480} \pm 0.265$ & $1.00\times$ & $0.92\times$ \\
128 & Flush & 16 & $148.64$ & $\mathbf{15.480} \pm 0.265$ & $1.00\times$ & $1.00\times$ \\
\rowcolor{bestrow}
128 & PACI  & 4  & $111.51$ & $15.632 \pm 0.009$ & $\mathbf{2.04\times}$ & $1.33\times$ \\
\rowcolor{bestrow}
128 & PACI  & 8  & $\mathbf{87.99}$  & $15.483 \pm 0.028$ & $1.84\times$ & $\mathbf{1.69\times}$ \\
\midrule
256 & Flush & 8  & $169.38$ & $15.350 \pm 0.089$ & $1.00\times$ & $0.73\times$ \\
256 & Flush & 16 & $124.27$ & $15.350 \pm 0.089$ & $1.00\times$ & $1.00\times$ \\
\rowcolor{bestrow}
256 & PACI  & 8  & $111.91$ & $15.399 \pm 0.042$ & $\mathbf{1.51\times}$ & $1.11\times$ \\
\rowcolor{bestrow}
256 & PACI  & 16 & $\mathbf{87.89}$ & $\mathbf{15.291} \pm 0.008$ & $1.41\times$ & $\mathbf{1.41\times}$ \\
\bottomrule
\end{tabular}
\end{table}

\subsection{Throughput scaling matches theory and \method{} removes bubble overhead}
\label{subsec:throughput_scaling}

\begin{takeawaybox}
\textbf{Key Takeaway.} 1F1B-flush scales according to pipeline bubble efficiency, while \method{} attains the corresponding fully utilized throughput with the same peak memory.
\end{takeawaybox}

\begin{figure}[t]
    \centering
    \includegraphics[width=0.9\linewidth]{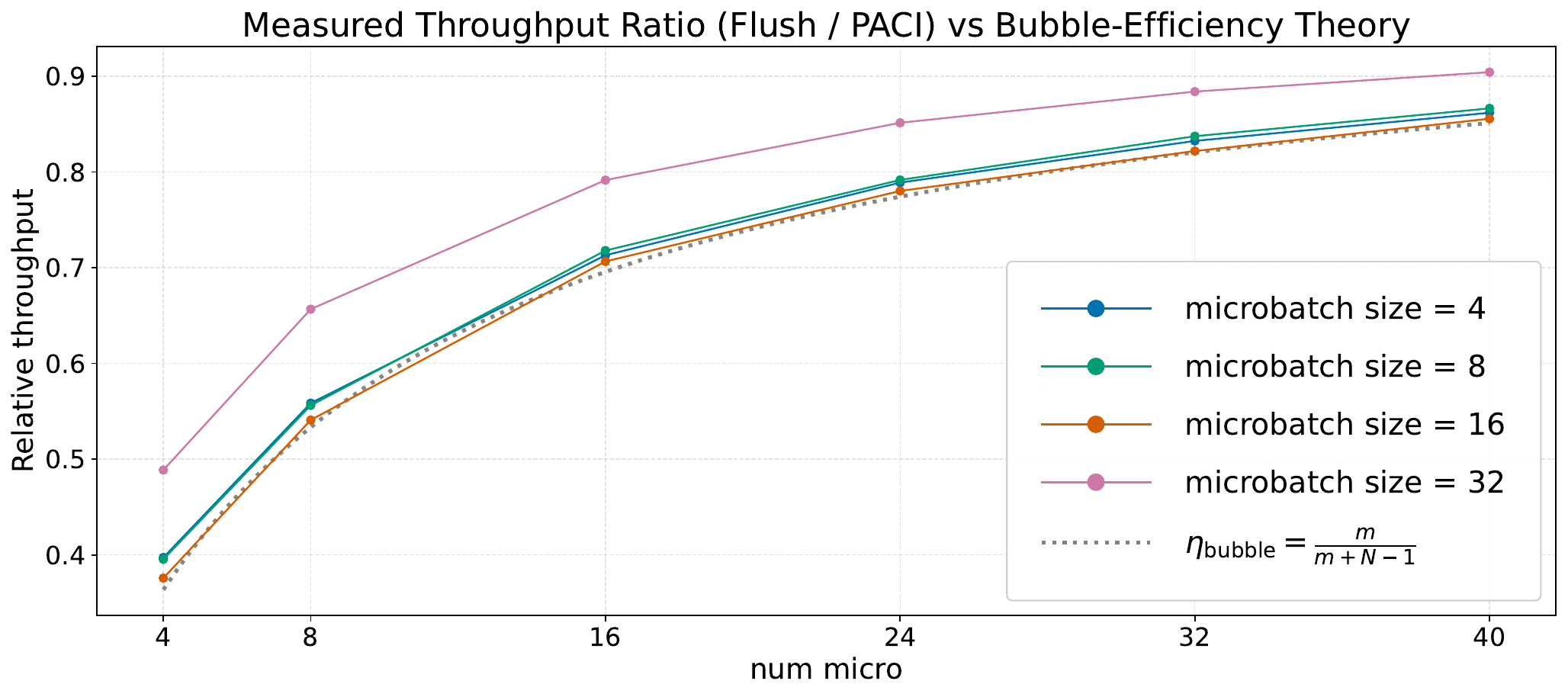}
    \caption{ 
Relative throughput of 1F1B-flush normalized by \method{}  as a function of the number of micro-batches $m$ for different micro-batch sizes.The empirical curves closely match the theoretical efficiency predicted by Eq.~\eqref{eq:bubble_efficiency_flush}, confirming that the throughput gap is explained by pipeline bubbles.
}
    \label{fig:microbatch_sweep_theory_vs_empirical_throughput}
\end{figure}

To answer Question~\textbf{Q3}, we sweep the number of micro-batches $m$ on GPT-2 Medium. Figure~\ref{fig:microbatch_sweep_theory_vs_empirical_throughput_micro_grid} shows the raw throughput curves: 1F1B-flush improves as $m$ increases, whereas \method{} remains nearly flat, consistent with bubble-free execution. When 1F1B-flush throughput is normalized by \method{}, the resulting curves  (Figure~\ref{fig:microbatch_sweep_theory_vs_empirical_throughput}) closely match the pipeline-efficiency prediction in Eq.~\eqref{eq:bubble_efficiency_flush}. This confirms that the measured throughput gap is explained by flush bubbles.
Figure~\ref{fig:max_mem_vs_num_micro} further shows that \method{} matches the peak memory of 1F1B-flush, except in the non-steady-state $m=4$ regime. Finally, Figure~\ref{fig:minibatch_sweep_flush_v1_measured_vs_bubble_prediction} shows why increasing $m$ is not a complete solution for synchronous schedules: bubble efficiency improves with $m$, but the corresponding decrease in micro-batch size eventually reduces kernel efficiency and lowers throughput. Thus, flush-based schedules must trade bubble reduction against kernel efficiency, while \method{} removes bubbles directly and uses micro-batching primarily to control version drift.

\section{Discussion}
\label{sec:discussion}

\method{} targets a simple operating point: the steady-state throughput of asynchronous 1F1B with the memory footprint of synchronous 1F1B-flush. This combines the main advantage of both regimes: no flush bubbles and no memory overhead. This sets up the central question; whether the resulting bounded forward/backward inconsistency preserves training stability and final quality. To show that, our experimental comparison focuses on synchronous 1F1B-flush. Flush provides the cleanest reference for standard training semantics and the same baseline memory footprint; replacing it with \method{} isolates the effect of trading synchronization for bounded version drift. More aggressive synchronous schedules mainly alter the systems trade-off by reducing bubbles through better scheduling, or additional memory. Prior asynchronous methods control inconsistency through weight stashing, prediction, or auxiliary state. Our experiments show \method{} matches synchronous training quality without the extra memory, prediction, or auxiliary computation other asynchronous methods require, while operating at fully utilized pipeline throughput. Activation checkpointing can also be combined with \method{} without extra parameter memory, although recomputation changes the inconsistency structure; we discuss this variant in Appendix~\ref{app:activation_checkpointing}. Combined with the efficiency-based large-scale comparison, these results suggest that bounded version drift can recover the utilization benefits of asynchronous execution without paying the memory or semantic costs of existing asynchronous consistency mechanisms.

\paragraph{Limitations}
Our evaluation is limited to GPT-2 Medium on OpenWebText, 8-stage pipelines, and a fixed set of training configurations; larger models, deeper pipelines, other datasets, modalities, optimizers, and schedules require further validation. We also leave activation-checkpointed \method{} to future work. Activation checkpointing can be combined with \method{} without extra parameter memory, but recomputation changes the inconsistency structure; see Appendix~\ref{app:activation_checkpointing}. Additionally, exact global gradient clipping requires synchronized full-model gradient norms and is incompatible with fully asynchronous execution; optimizer-level spike mitigation such as SPAM~\cite{huang2025spam} may provide a synchronization-free alternative. Second, we do not implement rollback for invalid gradients such as NaNs: if later stages have already updated, earlier stages may be unable to complete the corresponding backward pass. Whether explicit rollback is necessary, or whether partial stage updates can be tolerated, remains future work.
\bibliographystyle{plainnat}
\bibliography{egbib}

\appendix
\newpage
\section{Detailed comparison of pipeline parallelism methods}
\label{app:app_related}

\begin{table*}[h]
\centering
\caption{Detailed trade-offs in pipeline parallelism methods.}

\footnotesize
\setlength{\tabcolsep}{4pt}
\renewcommand{\arraystretch}{1.12}

\begin{tabular}{lccc}
\toprule
\textbf{Method} & \textbf{Extra Mem.} & \textbf{Bubble} & \textbf{F/B Incons.} \\
\midrule
\multicolumn{4}{l}{\textbf{Synchronous}} \\
\midrule
GPipe~\cite{huang2019gpipe} & +++ & High & 0 \\
PipeDream-Flush~\cite{narayanan2021memory} & 0 & High & 0 \\
DAPPLE~\cite{fan2021dapple} & 0 & High & 0 \\
Seq1F1B~\cite{sun2024seq1f1b} & 0 & Reduced & 0 \\
Megatron 1F1B-I~\cite{narayanan2021efficient} & +$^\star$& Reduced & 0 \\
Chimera~\cite{li2021chimera} & ++ & Reduced & 0 \\
ZB-2p~\cite{qi2023zero} & ++ & Near-0 & 0 \\
ZB-V~\cite{qi2023zero} & $0^{\dagger}$ & Near-0$^{\dagger}$ & 0 \\

\midrule
\multicolumn{4}{l}{\textbf{Asynchronous}} \\
\midrule
Na\"ive 1F1B \cite{harlap2018pipedream} & 0 & 0 & High \\
PipeDream~\cite{harlap2018pipedream} & +++ & 0 & $0^{\ddagger}$ \\
PipeDream-2BW~\cite{narayanan2021memory} & ++ & 0 & $0^{\ddagger}$ \\
AshPipe~\cite{hosoki2024ashpipe} & ++ & 0 & $0^{\ddagger,\ast}$ \\
PipeMare~\cite{yang2021pipemare} & ++ & 0 & High$^{\S}$ \\
SpecTrain~\cite{chen2018efficient} & ++ & 0 & Approx. \\
XPipe~\cite{guan2019xpipe} & ++ & 0 & Approx. \\
PipeOptim~\cite{guan2025pipeoptim} & +/++ & 0 & Approx. \\
AMDP~\cite{chenamdp} & +/++ & Near-0$^\times$& Low$^{\times}$ \\
\rowcolor{bestrow}
\textbf{\method{} (Ours)} & \textbf{0} & \textbf{0} & \textbf{low (Bounded)} \\
\bottomrule
\end{tabular}

\vspace{0.4em}
\begin{minipage}{0.96\linewidth}
\footnotesize
Extra Mem. is measured relative to na\"ive asynchronous 1F1B with the same model partitioning, micro-batch size, and activation-checkpointing policy. 
Here, $0$ denotes no additional memory beyond this baseline, while $+$, $++$, and $+++$ denote increasing extra activation, weight-version, or prediction-related memory. 
F/B Incons. denotes forward/backward weight-version inconsistency for the same micro-batch.

$^{\star}$ For 1F1B-I, $+$ refers to peak runtime memory overhead, e.g., activation and buffer residency from virtual pipeline stages, not additional parameter or optimizer state. This is consistent with the higher 1F1B-I peak memory reported by Zero Bubble~\cite[Table~4]{qi2023zero}.

$^{\dagger}$ZB-V attains near-zero bubbles with 1F1B-like peak memory only under its specialized V-shaped schedule and balanced timing assumptions.

$^{\ddagger}$Versioning eliminates F/B inconsistency, but delayed or stale-update semantics relative to fully synchronous training may remain.

$^{\S}$PipeMare uses learning-rate rescheduling and discrepancy correction to tolerate asynchronous delay and F/B mismatch; T2 adds a weight-sized velocity accumulator but does not enforce exact same-version F/B execution.

$^{\ast}$AshPipe uses stage-aware recomputation and version switching to reduce memory pressure caused by storing multiple weight versions; recomputation overhead is not counted as bubble overhead.

$^\times$ AMDP bounds mismatch through read-ahead restriction and multi-directional pipelines, but requires replicated logical stages and gradient synchronization across replicas increasing memory footprint.
\label{tab:pipeline_tradeoffs_detailed}

\end{minipage}

\end{table*}

\newpage
\section{Detailed experimental setup} \label{sec:appendix_experimental_setup}

\paragraph{Models and data.}
Our experiments are performed on GPT-2 Medium. All training was done from scratch on OpenWebText. The dataset is filtered to remove short documents (length $< 20$ words), split into 98\% training and 2\% validation prior to tokenization, and tokenized using the standard GPT-2 tokenizer. Tokens sequences are concatenated into blocks of length 1024 with eos inserted between each sequence in the block. Validation is performed every 5000 steps.

\paragraph{Optimization and training.}
We train all models using AdamW \cite{loshchilov2017decoupled} with $\beta_1 = 0.9$, $\beta_2 = 0.95$, and $\epsilon = 10^{-8}$. We use a peak learning rate of $\eta=3 \times 10^{-4}$ for batch size 256 and $\eta = 3\times\sqrt{\frac{1}{2}}\times10^{-4}$ for batch size 128 as suggested by~\cite{malladi2022sdes} to have comparable results  across batch sizes. We use a linear warmup \cite{vaswani2017attention} over 1\% of training steps followed by cosine decay \cite{loshchilov2016sgdr} to 10\% of the peak value. Weight decay is set to 0.1 (with no decay on bias and LayerNorm/embedding) and dropout to 0.1. All experiments are conducted in BF16 precision \cite{kalamkar2019study}.

\paragraph{Micro-batching.}
We use global batch sizes of 128 or 256 sequences. Training is performed using pipeline parallelism across 8 stages without data parallelism. Micro-batch size and accumulation factor $a$ vary across experiments. When the global batch size is not divisible by the number of micro-batches (Figure~\ref{fig:minibatch_sweep_flush_v1_measured_vs_bubble_prediction} only), we preserve the fixed global batch size by using heterogeneous micro-batch sizes: the first $B \bmod m$ micro-batches contain $\lfloor B/m \rfloor + 1$ sequences and the remaining micro-batches contain $\lfloor B/m \rfloor$ sequences. The accumulation factor controls the maximum forward/backward weight-version inconsistency according to Eq.~\eqref{eq:inconsistency_max}.

\paragraph{Pipeline configuration.}
We compare synchronous 1F1B-flush scheduling with our method, both implemented with pipeline parallelism across 8 GPUs. For the synchronous 1F1B-flush baseline, we fully train only the fastest micro-batch configuration for each global batch size. Under flush semantics, varying the number of micro-batches affects throughput but not the sequence of global optimizer updates, for a fixed global batch size, data order, and model partitioning. We therefore reuse the resulting validation trajectory across flush micro-batch counts and compute the corresponding runtimes from separately measured steady-state throughputs. All \method{} configurations are trained end-to-end, since their update trajectory depends on the induced weight-version inconsistency. Layers are partitioned using a custom strategy described in Appendix~\ref{app:partitioner} that balances compute and memory usage to maximize throughput under device constraints. Residual load imbalance is minimized but not entirely eliminated.

\paragraph{Systems and hardware.}
Experiments are conducted on a single node with 8 GPUs, primarily using NVIDIA RTX PRO 6000 Blackwell Max-Q GPUs. Some runs, used only in Section~\ref{subsec:training_stability}, were performed on a mixture of L40S, L40, A40, and A6000 Ada GPUs. GPUs are connected via PCIe. Experiments were run using Python 3.9, CUDA 12.8,  and a customized version of PyTorch 2.4 - described in Appendix ~\ref{sec:appendix_pytorch}.

\paragraph{Precision and memory.}
All experiments use BF16 precision without activation checkpointing or ZeRO/FSDP optimizations.

\paragraph{Activation checkpointing.}
\label{app:activation_checkpointing}

Our main experiments do not use activation checkpointing. However, \method{} is not fundamentally incompatible with checkpointing: discarded activations can be recomputed during the backward pass using the stage's current parameter version. This preserves the zero-extra-parameter-memory property of \method{}, since each stage still keeps only a single parameter copy and does not store old weight versions.
The caveat is that recomputation changes the form of forward/backward inconsistency. Without checkpointing, the backward pass differentiates activations produced by the original forward pass, but evaluates the gradient with respect to the current parameter version. With checkpointing, those activations are instead recomputed under the current parameter version before differentiation. Thus, the mismatch shifts from stored-old-activation/current-weight inconsistency to a mismatch between the downstream gradient, which was produced by the original pipeline computation, and local activations recomputed under newer weights. Although the same accumulation-based consistency bound still limits the number of parameter versions crossed, checkpointing changes how the mismatch manifests during backward computation; we leave the analysis and empirical evaluation of this variant to future work.

\paragraph{Memory measurement.}
We report peak GPU memory per device, measured using \texttt{nvidia-smi}, and report the maximum across all pipeline stages.

\paragraph{Time-to-accuracy.}
    We report training wall-clock time-to-accuracy, measured using steady-state training throughput excluding evaluation and I/O overheads. Token counts are measured from step 0. Accuracy is defined in terms of validation perplexity. We report the time required to reach thresholds of PPL $\leq$ 18, 17, and 16, corresponding to early, middle, and late training stages. The reported crossing time is determined by linearly interpolating between the two surrounding measurements before and after crossing.

\paragraph{Token budget and reproducibility.}
All methods are trained on a fixed token budget of 49.8B tokens (190K steps for batch size 256, 380K steps for batch size 128). All experiments use identical data ordering and are repeated across 3 random seeds; we report mean results with std as shaded bands.

\paragraph{Kernel and bubble efficiency.}
Bubble efficiency follows the standard analytical model of pipeline utilization. Kernel efficiency is not directly measured but inferred: we analyze deviations from theoretical throughput scaling when increasing the number of micro-batches while maintaining global batch-size which consequently reduces micro-batch size, attributing discrepancies to reduced kernel efficiency at smaller compute granularities.

\newpage

\section{Additional results}
\label{app:additional_results}
\begin{figure}[h]
    \centering
    \includegraphics[width=0.9\linewidth]{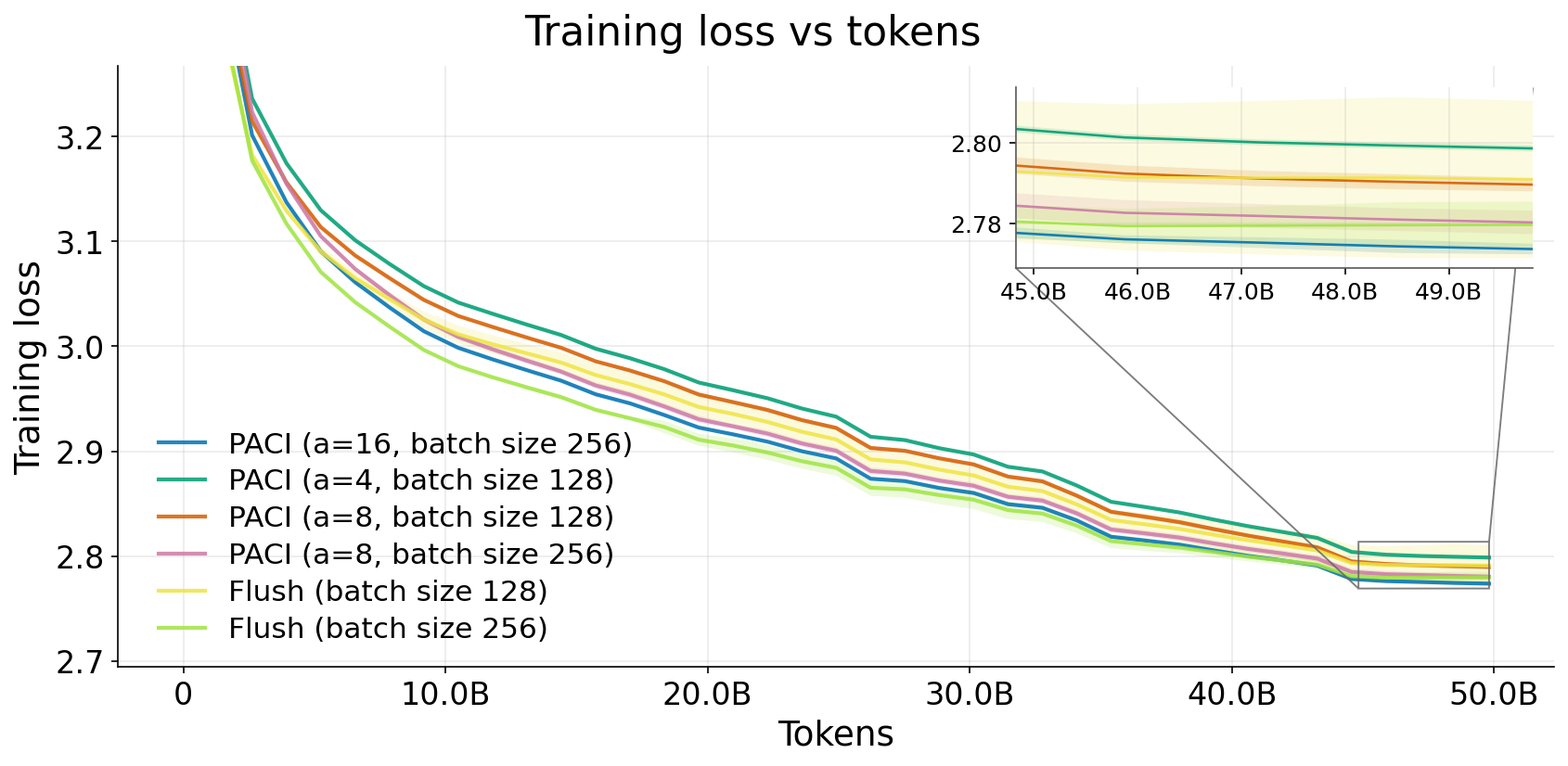}
    \caption{Training loss versus processed tokens for 1F1B-flush and \method{} under different accumulation factors. Bounded forward/backward inconsistency produces loss trajectories that closely track the synchronous baseline, with no evidence of instability or divergence.}
    \label{fig:training_loss_vs_tokens}
\end{figure}

\paragraph{Training dynamics.}
Figure~\ref{fig:training_loss_vs_tokens} shows the training loss as a function of processed tokens for both 1F1B-flush and \method{} under different accumulation factors. All configurations exhibit nearly identical convergence behavior, with \method{} closely tracking the synchronous baseline throughout training. This indicates that bounded forward/backward inconsistency does not introduce optimization instability or divergence.

\begin{figure}[h]
    \centering
    \includegraphics[width=0.9\linewidth]{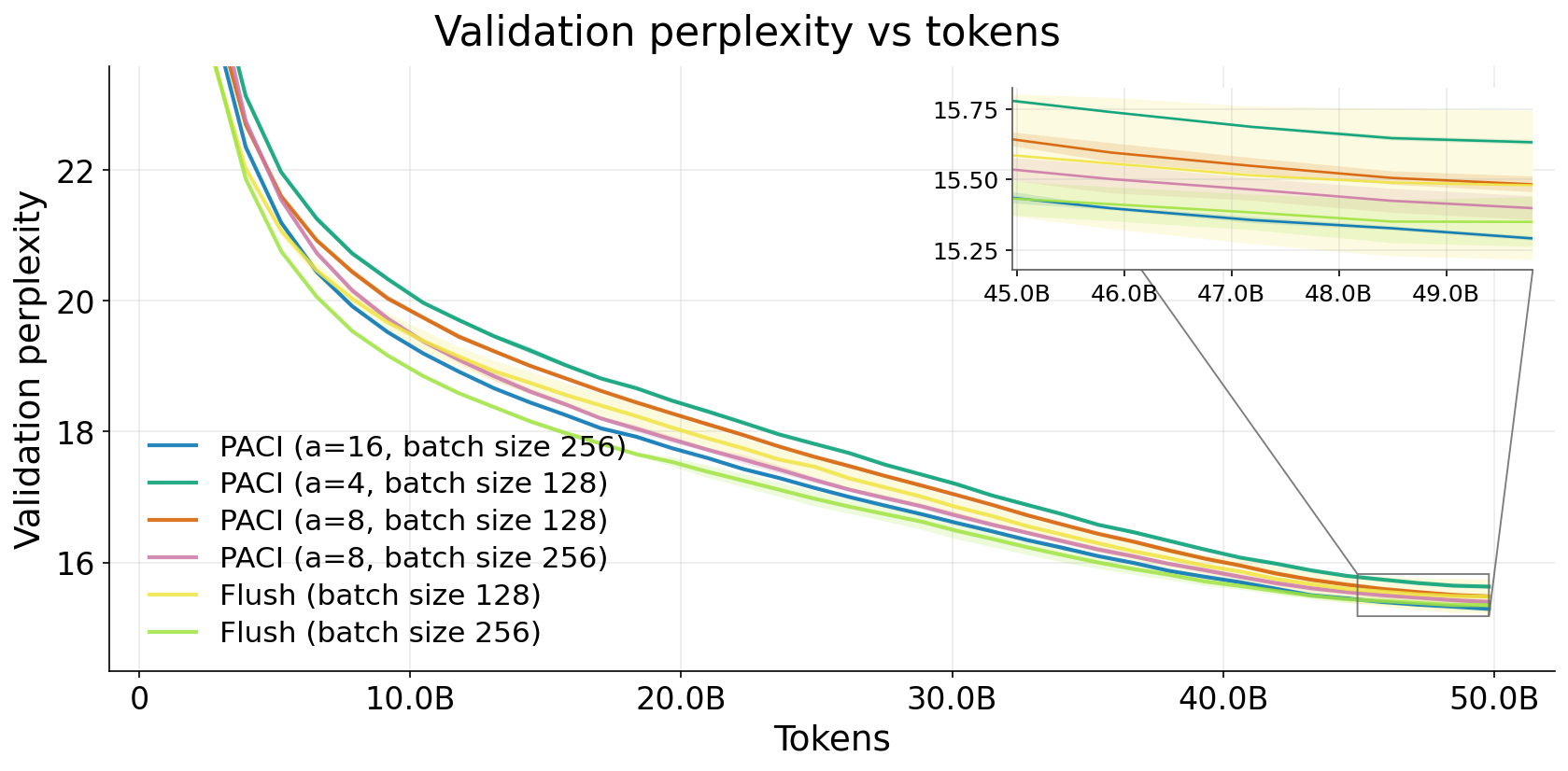}
    \caption{Validation perplexity versus processed tokens for 1F1B-flush and \method{} under different accumulation factors. The trajectories remain closely aligned across configurations, mirroring the training-loss behavior in Figure~\ref{fig:training_loss_vs_tokens} and indicating that bounded forward/backward inconsistency does not degrade generalization performance.}
    \label{fig:validation_perplexity_vs_tokens}
\end{figure}

\paragraph{Generalization performance.}
As shown in Figure~\ref{fig:validation_perplexity_vs_tokens}, validation perplexity remains consistent across all configurations. The close alignment between curves suggests that asynchronous execution with gradient accumulation preserves generalization performance, matching the behavior observed in training loss (Figure~\ref{fig:training_loss_vs_tokens}).

\begin{figure}[h]
    \centering
    \includegraphics[width=0.9\linewidth]{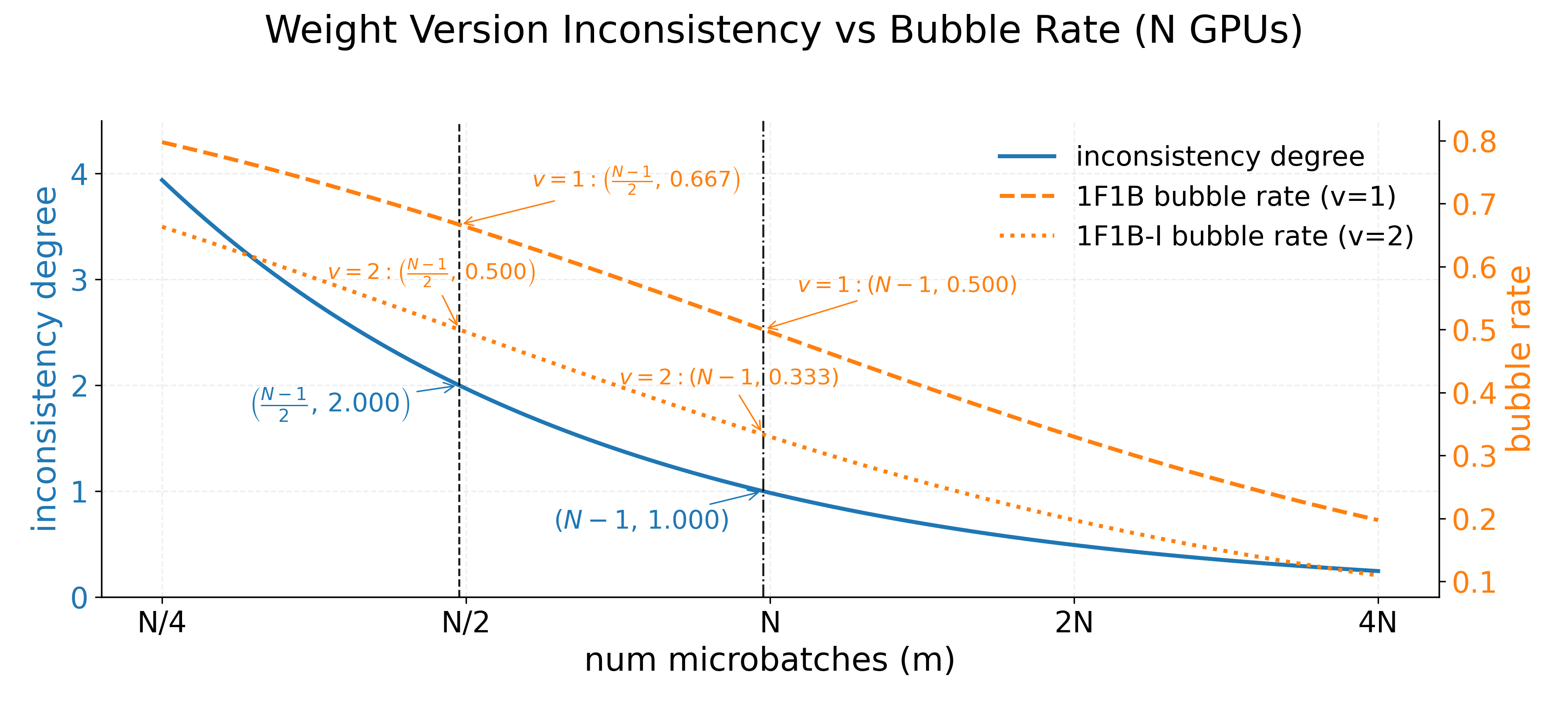}
    \caption{
    Bubble fraction versus forward/backward inconsistency as the number of micro-batches $m$ increases. 
    Synchronous 1F1B-flush and interleaved 1F1B use larger $m$ to amortize bubbles, whereas \method{} uses accumulation to reduce version drift. 
    Moderate $m$ is therefore sufficient for low inconsistency in \method{}, but not for high utilization in synchronous schedules.
    }
    \label{fig:inconsistency_vs_bubble}
\end{figure}

\begin{figure}[h]
    \centering
    \includegraphics[width=0.9\linewidth]{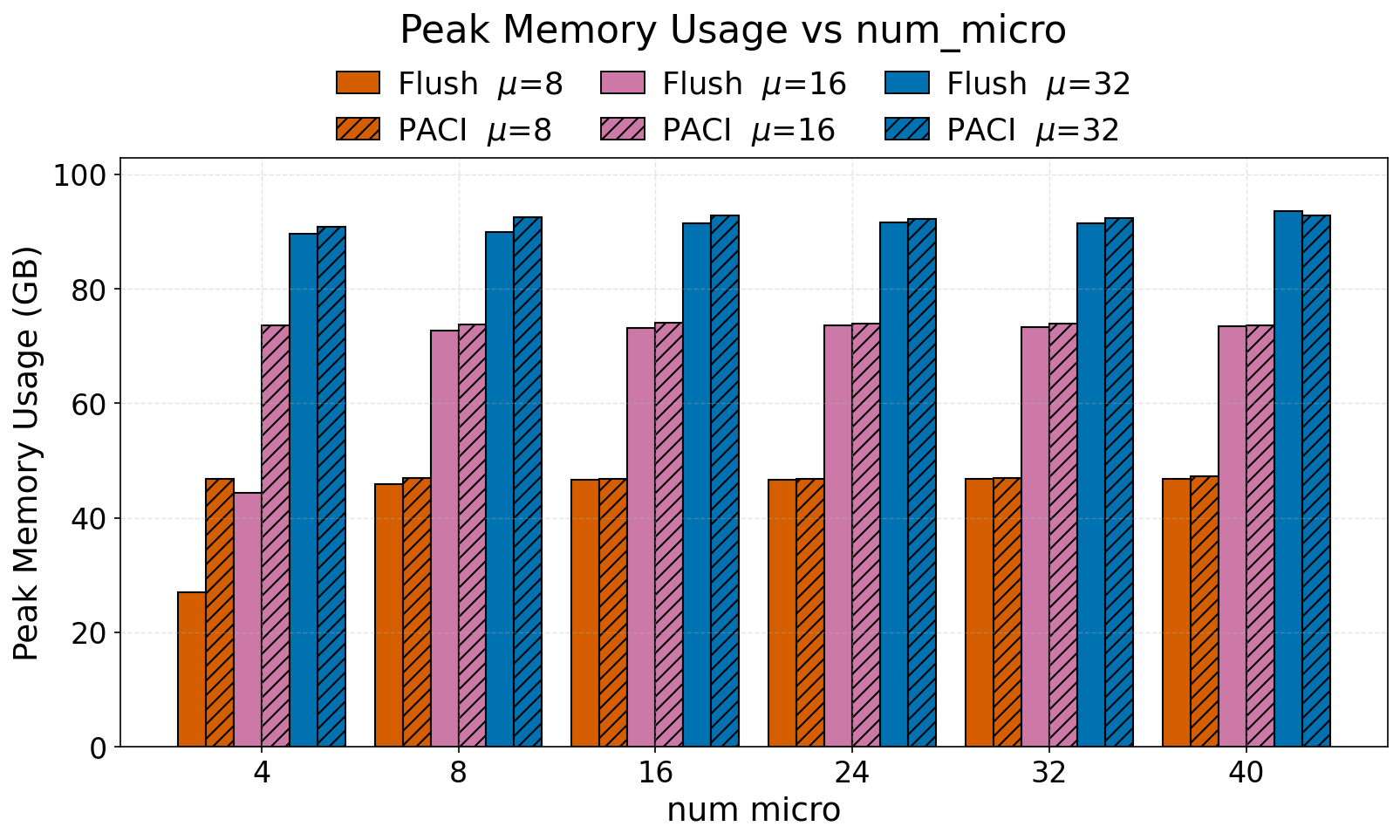}
    \caption{
Peak GPU memory usage for each micro-batch size ($\mu$) as a function of the number of micro-batches. 
Both methods exhibit identical memory usage across all steady-state configurations. The deviation at $m=4$ is due to under-utilization of the pipeline in 1F1B-flush, which prevents reaching steady-state execution.
}
    \label{fig:max_mem_vs_num_micro}
\end{figure}

\paragraph{Memory footprint.}
Figure~\ref{fig:max_mem_vs_num_micro} presents peak GPU memory usage as a function of the number of micro-batches $m$. Both methods exhibit identical memory consumption in all steady-state regimes, confirming that \method{} does not increase measured peak memory in these steady-state configurations. The deviation at small $m$ (e.g., $m=4$) is explained by pipeline under-utilization in 1F1B-flush, which prevents reaching steady-state execution.

\begin{figure}[h]
    \centering
    \includegraphics[width=0.9\linewidth]{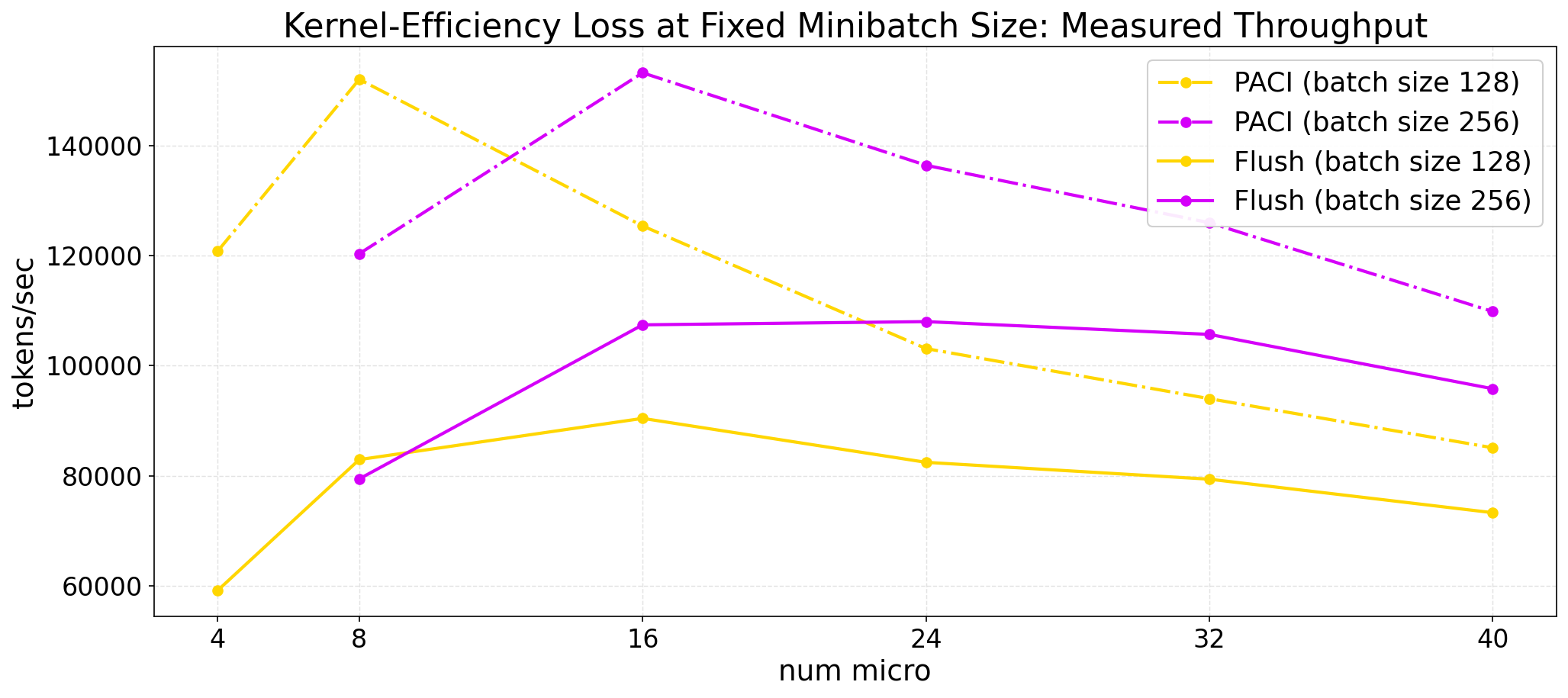}
    \caption{
Throughput as a function of the number of micro-batches $m$ for fixed mini-batch sizes $B \in \{128,256\}$. When $B$ is not divisible by $m$, we keep the global mini-batch size fixed and split it into heterogeneous micro-batches: the first $B \bmod m$ micro-batches have size $\lfloor B/m \rfloor + 1$, and the remaining micro-batches have size $\lfloor B/m \rfloor$. Increasing $m$ reduces the average micro-batch size, leading to a trade-off between improved pipeline utilization for 1F1B-flush and reduced kernel efficiency. \method{} removes bubble overhead and therefore primarily reflects kernel and communication effects, reaching peak performance at moderate $m$. Across all configurations, \method{} achieves higher throughput, with the gap narrowing at large $m$ where kernel inefficiency dominates.
}
    \label{fig:minibatch_sweep_flush_v1_measured_vs_bubble_prediction}
\end{figure}

\paragraph{Throughput trade-offs.}
Figure~\ref{fig:minibatch_sweep_flush_v1_measured_vs_bubble_prediction} illustrates throughput as a function of $m$ for fixed mini-batch sizes. Increasing $m$ improves flush pipeline utilization but reduces kernel efficiency due to smaller micro-batches. \method{} consistently achieves higher throughput by eliminating pipeline bubbles, with the performance gap narrowing at large $m$ where kernel inefficiency dominates.

\begin{table*}[h] 
\centering
\footnotesize
\begin{tabular}{cccrrcc}
\toprule
Batch & PPL threshold & Method & Num. micro & Runtime & Speedup$_{\mathrm{match}}$ & Speedup$_{\mathrm{best}}$ \\
\midrule
128 & $\leq 18$ & Flush & 4  & $88.39 \pm 6.48$h  & $1.00\times$ & $0.65\times$ \\
128 & $\leq 18$ & Flush & 8  & $63.03 \pm 4.62$h  & $1.00\times$ & $0.92\times$ \\
128 & $\leq 18$ & Flush & 16 & $57.72 \pm 4.23$h  & $1.00\times$ & $1.00\times$ \\
128 & $\leq 18$ & \method{}  & 4  & $50.57 \pm 0.17$h  & $\mathbf{1.75\times}$ & $1.14\times$ \\
128 & $\leq 18$ & \method{}  & 8  & $\mathbf{37.32} \pm 0.39$h  & $1.69\times$ & $\mathbf{1.55\times}$ \\
\midrule
128 & $\leq 17$ & Flush & 4  & $129.08 \pm 5.47$h & $1.00\times$ & $0.65\times$ \\
128 & $\leq 17$ & Flush & 8  & $92.05 \pm 3.90$h  & $1.00\times$ & $0.92\times$ \\
128 & $\leq 17$ & Flush & 16 & $84.29 \pm 3.57$h  & $1.00\times$ & $1.00\times$ \\
128 & $\leq 17$ & \method{}  & 4  & $69.89 \pm 0.36$h  & $\mathbf{1.85\times}$ & $1.21\times$ \\
128 & $\leq 17$ & \method{}  & 8  & $\mathbf{52.86} \pm 0.23$h  & $1.74\times$ & $\mathbf{1.59\times}$ \\
\midrule
128 & $\leq 16$ & Flush & 4  & $177.07 \pm 10.34$h & $1.00\times$ & $0.65\times$ \\
128 & $\leq 16$ & Flush & 8  & $126.28 \pm 7.38$h  & $1.00\times$ & $0.92\times$ \\
128 & $\leq 16$ & Flush & 16 & $115.63 \pm 6.75$h  & $1.00\times$ & $1.00\times$ \\
128 & $\leq 16$ & \method{}  & 4  & $92.92 \pm 0.38$h   & $\mathbf{1.91\times}$ & $1.24\times$ \\
128 & $\leq 16$ & \method{}  & 8  & $\mathbf{70.31} \pm 0.41$h   & $1.80\times$ & $\mathbf{1.64\times}$ \\
\midrule
256 & $\leq 18$ & Flush & 8  & $49.85 \pm 0.92$h  & $1.00\times$ & $0.73\times$ \\
256 & $\leq 18$ & Flush & 16 & $36.57 \pm 0.68$h  & $1.00\times$ & $1.00\times$ \\
256 & $\leq 18$ & \method{}  & 8  & $40.16 \pm 0.92$h  & $\mathbf{1.24\times}$ & $0.91\times$ \\
256 & $\leq 18$ & \method{}  & 16 & $\mathbf{29.47} \pm 0.41$h  & $\mathbf{1.24\times}$ & $\mathbf{1.24\times}$ \\
\midrule
256 & $\leq 17$ & Flush & 8  & $81.77 \pm 3.84$h  & $1.00\times$ & $0.73\times$ \\
256 & $\leq 17$ & Flush & 16 & $59.99 \pm 2.82$h  & $1.00\times$ & $1.00\times$ \\
256 & $\leq 17$ & \method{}  & 8  & $60.18 \pm 0.88$h  & $\mathbf{1.36\times}$ & $1.00\times$ \\
256 & $\leq 17$ & \method{}  & 16 & $\mathbf{45.15} \pm 0.34$h  & $1.33\times$ & $\mathbf{1.33\times}$ \\
\midrule
256 & $\leq 16$ & Flush & 8  & $119.02 \pm 4.13$h & $1.00\times$ & $0.73\times$ \\
256 & $\leq 16$ & Flush & 16 & $87.32 \pm 3.03$h  & $1.00\times$ & $1.00\times$ \\
256 & $\leq 16$ & \method{}  & 8  & $84.11 \pm 0.78$h  & $\mathbf{1.42\times}$ & $1.04\times$ \\
256 & $\leq 16$ & \method{}  & 16 & $\mathbf{64.00} \pm 0.11$h  & $1.36\times$ & $\mathbf{1.36\times}$ \\
\bottomrule
\end{tabular}
\caption{
Runtime to reach each PPL threshold. 
Speedup$_{\mathrm{match}}$ compares \method{} against flush with the same number of micro-batches.
Speedup$_{\mathrm{best}}$ compares each configuration against the fastest flush baseline for the same batch size and PPL threshold. 
}
\label{tab:time_to_accuracy}

\end{table*}

\begin{figure}[h]
    \centering
    \includegraphics[width=0.9\linewidth]{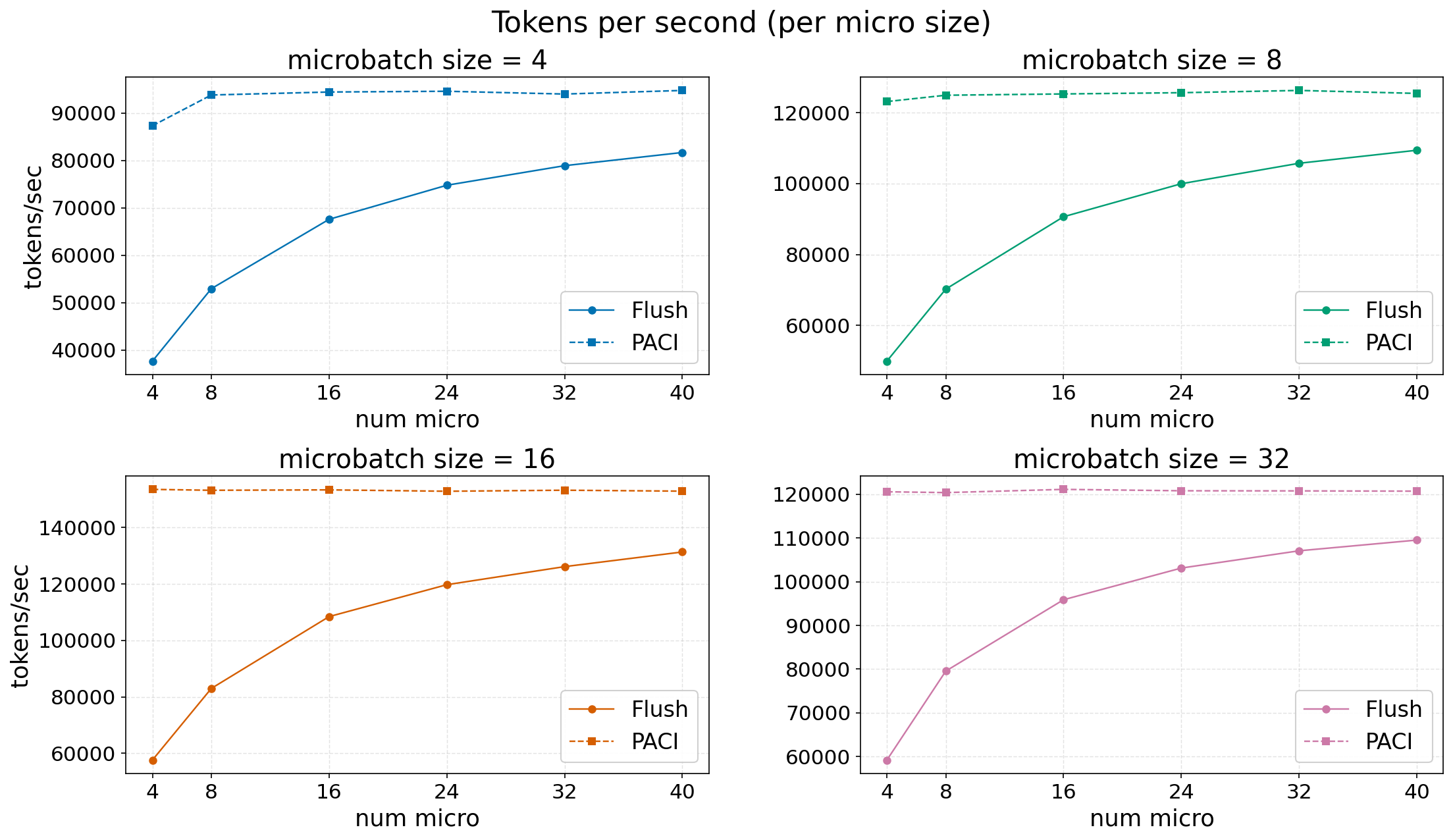}
    \caption{Throughput as a function of the number of micro-batches for 1F1B-flush and \method{}. \method{} provides the largest throughput gains at low-to-moderate micro-batch counts, where pipeline bubbles dominate execution time. As the number of micro-batches increases, the flush baseline becomes increasingly efficient and asymptotically approaches the throughput of \method{}.}
    \label{fig:microbatch_sweep_theory_vs_empirical_throughput_micro_grid}
\end{figure}

\paragraph{Micro-batch efficiency analysis.}
A more detailed breakdown is provided in Figure~\ref{fig:microbatch_sweep_theory_vs_empirical_throughput_micro_grid}, which reports throughput across different micro-batch sizes. The advantage of \method{} is most pronounced at low-to-moderate number of micro-batches $m$, where bubble overhead is significant. As $m$ increases, flush converges to \method{} due to increased pipeline efficiency while only reaching it as $m \rightarrow \infty $.

\newpage
\paragraph{Time-to-accuracy.}
Table~\ref{tab:time_to_accuracy} reports wall-clock time to reach validation perplexity thresholds 18, 17, and 16. Across all thresholds and batch sizes, \method{} reaches the target perplexity faster than 1F1B-flush at the same number of micro-batches, with speedup increasing as the target becomes stricter. For batch size 128, the matched speedup grows from $1.75\times$ at PPL $\leq 18$ to $1.91\times$ at PPL $\leq 16$ for $m=4$, and from $1.69\times$ to $1.80\times$ for $m=8$. A similar trend appears at batch size 256, where the matched speedup increases from $1.24\times$ to up to $1.42\times$.
This trend reflects the token-level behavior in Figures~\ref{fig:training_loss_vs_tokens} and~\ref{fig:validation_perplexity_vs_tokens}: the gap between \method{} and 1F1B-flush narrows as training progresses, suggesting that the effect of bounded inconsistency diminishes over time. As a result, the wall-clock benefit increasingly reflects the underlying throughput advantage of bubble-free execution. Even compared with the fastest flush configuration for each threshold, \method{} remains faster in the later regime, reaching up to $1.64\times$ Speedup$_{\mathrm{best}}$ at batch size 128 and $1.36\times$ at batch size 256. Thus, \method{} converts its throughput advantage into faster time-to-accuracy, with gains becoming stronger as training stabilizes.

\section{Additional method details}
\label{sec:appendix_method_details}

\subsection{Pipeline setup and notation}

We consider training a model partitioned across $N$ pipeline stages. Stage $i \in \{1,\ldots,N\}$ has parameters $\theta_i$ and computes a function $F_i$, so the full model is
\begin{equation}
F(x;\theta)
=
F_N \circ F_{N-1} \circ \cdots \circ F_1(x),
\end{equation}
where $\theta=\{\theta_1,\ldots,\theta_N\}$. Let $\theta_i^{(t)}$ denote the parameters of stage $i$ after its $t$-th local optimizer update.

For micro-batch $m$, the forward activations satisfy
\begin{equation}
h_{m,0}=x_m,
\qquad
h_{m,i}
=
F_i(h_{m,i-1};\theta_i^{(t_{m,i}^{F})}),
\end{equation}
where $t_{m,i}^{F}$ is the local update index at stage $i$ when the forward pass of micro-batch $m$ is computed. In asynchronous 1F1B, the corresponding backward pass may arrive after additional local updates. Let $t_{m,i}^{B}$ be the local update index used during backward. The forward/backward inconsistency is
\begin{equation}
\Delta_{m,i}
=
t_{m,i}^{B} - t_{m,i}^{F}.
\end{equation}
In the main text, we write $\Delta_i$ when the micro-batch is clear from context.

\subsection{Inconsistency bound}

For each stage $i$, \method{} enforces a local unresolved-forward invariant. Let $q_i$ denote the number of micro-batches whose forward pass has completed at stage $i$ but whose corresponding backward pass has not yet returned to stage $i$. The stage increments $q_i$ after a local forward pass and decrements
$q_i$ after the corresponding local backward pass. A new forward pass is admitted only if, before admitting it,
\begin{equation}
q_i < N+1-i .
\end{equation}
Under the one-indexed stage convention used in this paper, $N-i$ is the downstream pipeline depth of stage $i$. Since $q_i$ is integer-valued, the admission rule is equivalent to requiring $q_i \le N-i$ before the new forward is issued. Thus, when a micro-batch $m$ is admitted at stage $i$, at most $N-i$ earlier forward passes at that stage remain unresolved. After admitting $m$, the total unresolved-forward count, including $m$, is bounded by $N+1-i$.
Therefore, at most $N-i$ local backward passes can complete between the forward computation of micro-batch $m$ at stage $i$ and its corresponding backward computation. Since \method{} applies one optimizer update only after every $a$ local backward passes, the number of parameter versions crossed by this micro-batch satisfies
\begin{equation}
\Delta_{m,i}
\le
\left\lceil \frac{N-i}{a} \right\rceil .
\label{eq:appendix_inconsistency_bound}
\end{equation}
The maximum occurs at the first stage, yielding
\begin{equation}
\Delta_{\max}
=
\max_{m,i} \Delta_{m,i}
\le
\left\lceil \frac{N-1}{a} \right\rceil .
\end{equation}

Thus, the accumulation factor $a$ controls parameter-version drift under an explicit scheduler invariant, without relying on balanced stage times. Increasing $a$ slows the evolution of local parameter versions and decreases the number of versions crossed between forward and backward computation.

\subsection{Stage-level execution rule}

Algorithm~\ref{alg:method_stage_execution} gives the local event rule for each stage. The pipeline remains asynchronous: stages execute forward and backward computations independently as soon as the required input tensors arrive. The main modification relative to standard asynchronous 1F1B is that optimizer updates are delayed until $a$ local gradients have been accumulated.

\begin{algorithm}[H]
\caption{Stage-$i$ execution in \method{}}
\label{alg:method_stage_execution}
\begin{algorithmic}[1]
\State Initialize gradient accumulator $G_i \leftarrow 0$
\State Initialize local backward counter $c_i \leftarrow 0$
\State Initialize unresolved-forward counter $q_i \leftarrow 0$
\For{each ready event at stage $i$}
\If{forward input for micro-batch $m$ is available and $q_i \leq N-i$}
        \State Compute $h_{m,i} \leftarrow F_i(h_{m,i-1}; \theta_i)$
        \State Store activations required for backpropagation
        \State Send $h_{m,i}$ to stage $i+1$
        \State $q_i \leftarrow q_i + 1$
    \EndIf
    \If{backward input for micro-batch $m$ is available}
        \State Compute local gradient $g_{m,i}$ and input gradient $\nabla h_{m,i-1}$
        \State Accumulate $G_i \leftarrow G_i + g_{m,i}$
        \State Send $\nabla h_{m,i-1}$ to stage $i-1$
        \State $c_i \leftarrow c_i + 1$
        \State $q_i \leftarrow q_i - 1$
        \If{$c_i = a$}
            \State Update $\theta_i$ using accumulated gradient $G_i$
            \State Reset $G_i \leftarrow 0$, $c_i \leftarrow 0$
        \EndIf
    \EndIf
\EndFor
\end{algorithmic}
\end{algorithm}

\subsection{Memory and synchronization properties}

\method{} stores one parameter copy per stage. It does not require weight stashing, future-weight prediction, or global synchronization between stages. Each stage maintains only the usual activations needed for backpropagation and accumulates the gradients into the usual gradient buffer. Consequently, the method introduces no additional weight-memory overhead relative to asynchronous 1F1B with the same model partitioning.
\method{} also preserves asynchronous execution. Parameter updates are local to each stage and occur after a fixed number of local backward passes. No global flush or cross-stage barrier is introduced. This is what allows \method{} to retain the throughput behavior of a bubble-free pipeline while bounding forward/backward inconsistency.

\subsection{Memory-constrained pipeline partitioning}
\label{app:partitioner}

We split a sequential model $f = f_L \circ f_{L-1} \circ \cdots \circ f_1$ across $N$ pipeline stages (devices) by choosing $N{-}1$ cut points $0 = c_0 < c_1 < \cdots < c_{N-1} < c_N = L$ that induce a contiguous partition $G_j = \{c_{j-1}{+}1, \dots, c_j\}$. The choice of cut points trades two competing objectives:

\begin{enumerate}
    \item \textbf{Throughput.} Pipeline throughput is bottle-necked by the slowest stage; we therefore want to minimize the maximum stage time $\max_j T(G_j)$, where $T(G_j)$ is the sum of the per-layer forward times in $G_j$.
    \item \textbf{Per-device memory.} To approximate the steady state memory footprint of a particular split, we split the per-layer footprint into a \emph{static} component $s_\ell$ - parameters, gradients and optimizer state,  which are resident on the stage regardless of pipeline occupancy --- and a per-micro-batch \emph{activation} component $a_\ell$, the bytes saved by autograd for the backward pass on a single micro-batch. In a 1F1B steady state, stage $j$ must keep activations alive for $w_j = N - j + 1$ in-flight microbatches simultaneously (the first stage carries $N$ microbatches, the last carries one)\footnote{Here, $w_j$ counts the total number of in-flight micro-batches simultaneously occupying memory. This differs from $q_i$ in the paper's bound in Eq.~\eqref{eq:inconsistency_max}, which counts only the unresolved forward passes strictly preceding the admission of the current micro-batch.}, while the static component is paid only once. The peak memory of stage $j$ is therefore 
\end{enumerate}
\begin{equation}    
M_j(G_j) \;=\; \underbrace{\sum_{\ell \in G_j} s_\ell}_{\text{static}}
\;+\; w_j\!\!\underbrace{\sum_{\ell \in G_j} a_\ell}_{\text{activations}} .
\end{equation}

We begin by profiling $t_\ell$, $s_\ell$ and $a_\ell$ once on a single device. Forward times $t_\ell$ are obtained by running $R$ forward passes of layer $\ell$ on a micro-batch and measuring wall-clock time after CUDA synchronization. Empirically, on the transformer-based architectures we study using the transformer blocks as our most granular groupings, per-layer backward time scales proportionally to forward time, so $t_\ell$ alone is a faithful proxy for total per-layer compute - the proportionality constant is stage-invariant and therefore does not affect the argmin in~\eqref{eq:partitioner}. Because the final layer computes the loss and the embedding layer is empirically always grouped with following transformer blocks, inter-layer boundary activations are roughly uniform in size; thus, cut placement has negligible effect on inter-stage communication volume. We therefore omit the communication term from the constraint accordingly. The static footprint $s_\ell$ is the bytes occupied by the layer's parameters, gradients and optimizer state after one fwd/bwd/optimizer step; the activation footprint $a_\ell$ is the bytes saved by autograd for the backward pass on a single micro-batch, tracked through PyTorch's \textit{saved\_tensors\_hooks} mechanism.

\paragraph{Optimization problem.}
Given a per-device memory budget $B$, we seek a partition that solves
\begin{equation}
\label{eq:partitioner}
\begin{aligned}
\min_{\,c_0,\dots,c_N\,} \quad & \max_{j \in [N]} T(G_j) \\
\text{s.t.}\quad & \sum_{\ell \in G_j} s_\ell + w_j \sum_{\ell \in G_j} a_\ell
\;\le\; B \qquad \forall\, j \in [N], \\
& 0 = c_0 < c_1 < \cdots < c_N = L .
\end{aligned}
\end{equation}
Problem~\eqref{eq:partitioner} is a \emph{constrained min-max
linear partitioning} problem augmented with a stage-dependent
memory feasibility constraint that captures the 1F1B in-flight
micro-batch footprint.

\paragraph{Algorithm.}
We solve~\eqref{eq:partitioner} with a two-pass dynamic
program. Let
\begin{equation}
T(a,b) \;=\; \sum_{\ell=a+1}^{b} t_\ell, \qquad
M_j(a,b) \;=\; \sum_{\ell=a+1}^{b} s_\ell \;+\; w_j \sum_{\ell=a+1}^{b} a_\ell ,
\end{equation}
We define $D[i][j]$ as the minimum achievable max-stage-time when the first $i$ layers are placed into the first $j$ stages while every
stage respects the memory budget. The recursion is
\begin{equation}
\label{eq:dp}
D[i][j] \;=\; \min_{\substack{j-1 \le x < i \\ M_j(x,i) \,\le\, B}}
\max\bigl( D[x][j-1], T(x,i) \bigr),
\end{equation}
with base case $D[i][1] = T(0,i)$ if $M_1(0,i) \le B$ and $+\infty$
otherwise. The optimal cut points are recovered by backtracking
through the argmin table $\pi[i][j]$.

\paragraph{Infeasibility fallback.}
If the budget $B$ is too tight,~\eqref{eq:partitioner} may be
infeasible ($D[L][N] = +\infty$). Rather than failing, we run a
second pass that operates over pairs $(o, t)$ ordered lexicographically, where $o = \max(0, M_j(G_j) - B)$
is the per-stage overshoot. The fallback minimizes the worst
overshoot first and uses the maximum stage time as a tie-breaker, so the returned partition is the most memory-balanced configuration that still respects the topology constraint. In practice, the first pass succeeds for every experiment we report and the fallback is only invoked for diagnostic sweeps.

\subsection{PyTorch modifications for asynchronous execution} 
\label{sec:appendix_pytorch}

Our implementation is based on PyTorch 2.4.0, with one modification to the version-counter check in order to enable backward execution after local parameter updates have occurred. The modification does not alter backward kernels; it only permits the asynchronous version mismatch that \method{} intentionally studies. Code for the PyTorch customization and the complete training implementation is available in the GitHub repository.

\paragraph{Background.}
Recent versions of PyTorch include a tensor versioning mechanism that tracks in-place updates to parameters. This mechanism is used by autograd to detect situations where gradients are computed using tensors that have been modified since their use in the forward pass. In such cases, PyTorch raises an error to prevent inconsistent gradient computations.
While this behavior is desirable for standard synchronous training, it prevents execution models in which forward and backward passes may intentionally operate on slightly different parameter versions, as is the case in \method{}.

\paragraph{Modification.}
To enable this execution model, we introduce a single additional control flag, \texttt{freeze\_version\_update}, which disables version counter increments during parameter updates. Concretely, the primary change to the PyTorch codebase is in the version counter update logic and related python wrappers:

\begin{verbatim}
void bump() {
  TORCH_CHECK(
      version_counter_ || InferenceMode::is_enabled(),
      "...");
  if (version_counter_ && (!freeze_version_update)) {
    ++version_counter_->version_;
  }
}
\end{verbatim}

This flag is exposed to Python via a lightweight wrapper, allowing us to selectively disable version tracking during training.
Disabling version counter updates prevents autograd from raising errors when gradients are computed using parameters that have been updated since the forward pass. Importantly, this modification does not alter gradient computation itself, but only removes the runtime consistency check.

\end{document}